\documentclass{article}

\usepackage[preprint]{neurips_2020}

\usepackage[utf8]{inputenc} 
\usepackage[T1]{fontenc}    
\usepackage{hyperref}       
\usepackage{url}            
\usepackage{booktabs}       
\usepackage{amsfonts}       
\usepackage{nicefrac}       
\usepackage{microtype}      
\usepackage{amsmath}
\usepackage{amssymb}
\usepackage{color}
\usepackage{graphicx}
\usepackage{caption}
\usepackage{subcaption}
\usepackage{soul}
\usepackage{todonotes}
\usepackage[ruled,vlined]{algorithm2e}
\usepackage{algorithmic}

\newcommand{\pd}[2]{\frac{\partial {#1}}{\partial {#2}}}

\newcommand{\mc}[0]{\mathcal}
\newcommand{\vc}[0]{\boldsymbol}
\newcommand{\mtx}[0]{\mathbf}

\newcommand{\reals}[0]{\mathbb R}
\newcommand{\doubleE}[0]{\mathbb E}

\newcommand{\scriptB}[0]{\mc B}

\newcommand{\scriptF}[0]{\mc F}
\newcommand{\scriptG}[0]{\mc G}

\newcommand{\scriptN}[0]{\mc N}
\newcommand{\scriptO}[0]{\mc O}
\newcommand{\scriptP}[0]{\mc P}

\newcommand{\scriptU}[0]{\mc U}

\newcommand{\GP}[2]{\scriptG \scriptP \left({#1}, {#2} \right)}

\DeclareMathOperator*{\argmin}{argmin} 

\setcitestyle{numbers}

\title{Bayesian Hidden Physics Models: Uncertainty Quantification for Discovery of Nonlinear Partial Differential Operators from Data}

\author{%
  Steven Atkinson \\
  GE Research \\
  \texttt{steven.atkinson1@ge.com}
}

\begin{document}

\maketitle

\begin{abstract}
    What do data tell us about physics---and what \textit{don't} they tell us?
    There has been a surge of interest in using machine learning models to discover governing physical laws such as differential equations from data, but current methods lack uncertainty quantification to communicate their credibility.
    This work addresses this shortcoming from a Bayesian perspective.
    We introduce a novel model comprising ``leaf'' modules that learn to represent distinct experiments' spatiotemporal functional data as neural networks and a single ``root'' module that expresses a nonparametric distribution over their governing nonlinear differential operator as a Gaussian process.
    Automatic differentiation is used to compute the required partial derivatives from the leaf functions as inputs to the root.
    Our approach quantifies the reliability of the learned physics in terms of a posterior distribution over operators and propagates this uncertainty to solutions of novel initial-boundary value problem instances.
    Numerical experiments demonstrate the method on several nonlinear PDEs.
\end{abstract}

\section{Introduction}
An enticing promise of modern machine learning applied to scientific domains is the discovery of novel physics from data.
Discovery of differential equations from data \cite{ramsay1996principal, bongard2007automated, rudy2017data} has featured prominently particular goal under this larger effort.
While earlier methods relied on symbolic manipulation of simple function models such as polynomials \cite{ramsay1996principal, rudy2017data} or using numerical finite difference schemes \cite{bongard2007automated} to extract the required differentials from measurements, a more contemporary approach leveraging automatic differentiation within ML frameworks \cite{abadi2016tensorflow, paszke2019pytorch, bradbury2018jax} and highly-effective methods for high-dimensional gradient-based stochastic optimization \cite{duchi2011adaptive, hinton2012neural, zeiler2012adadelta, sutskever2013importance, kingma2014adam} has opened up the possibility of using highly expressive function models such as neural networks to represent solution to a differential equation directly (as so-called ``physics-informed neural networks'' \cite{raissi2019physics}) or other targeted unknown elements within the larger analysis of differential equations such as the derivative operator in ordinary or partial differential equations \cite{chen2018neural, raissi2018deep}.

A key shortcoming of existing methods is that it is fundamentally unclear what limitations exist concerning knowledge about the governing differential equations arising from the finite nature of the data driving learning.
While promising results have shown that black-box methods for learning differential operators can extrapolate to unseen initial-boundary conditions \cite{raissi2018deep}, existing methods can fail without warning due to a lack of any formal uncertainty quantification (UQ) over the learned physics.
The only current recourse is validation of every prediction against the ground truth physics, defeating the purpose of predictive learning.

\begin{figure}[hbt]
    \centering
    \begin{subfigure}[b]{0.58\textwidth}
         \centering
         \includegraphics[width=\textwidth]{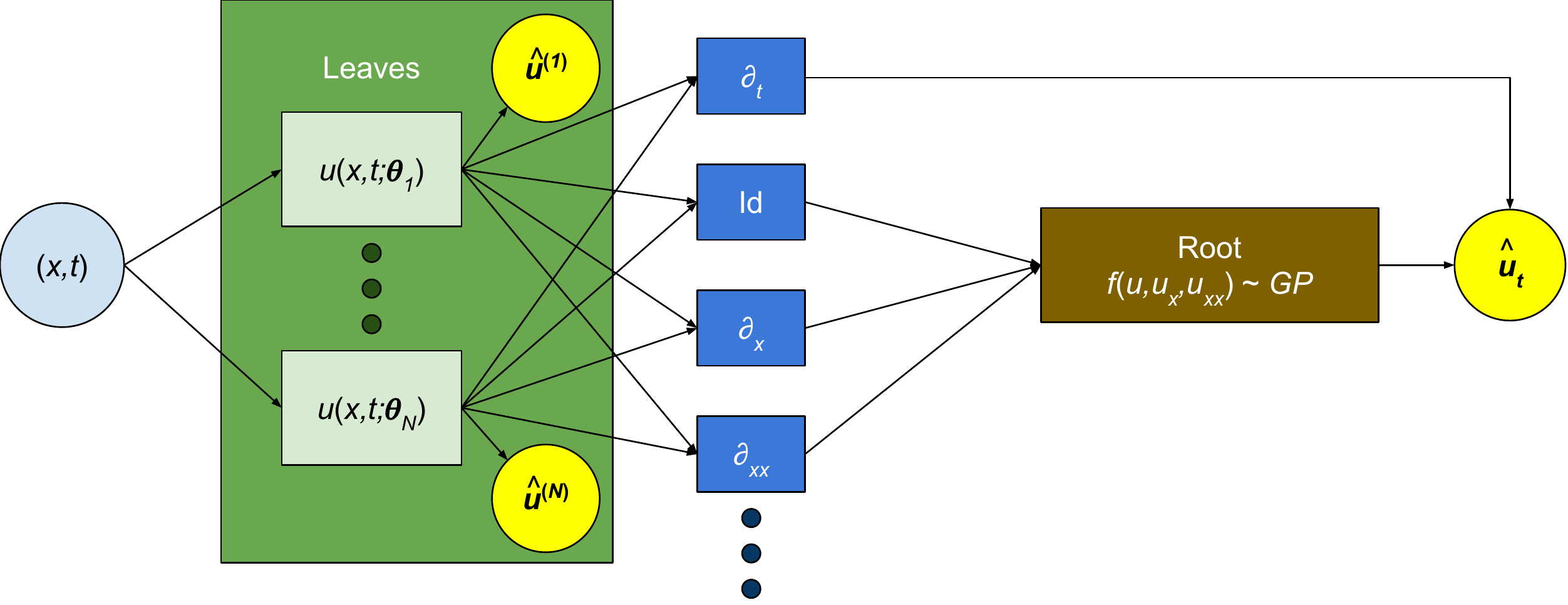}
         \caption{Bayesian HPM}
         \label{fig:BHPM}
    \end{subfigure}
    \begin{subfigure}[b]{0.38\textwidth}
         \centering
         \includegraphics[width=\textwidth]{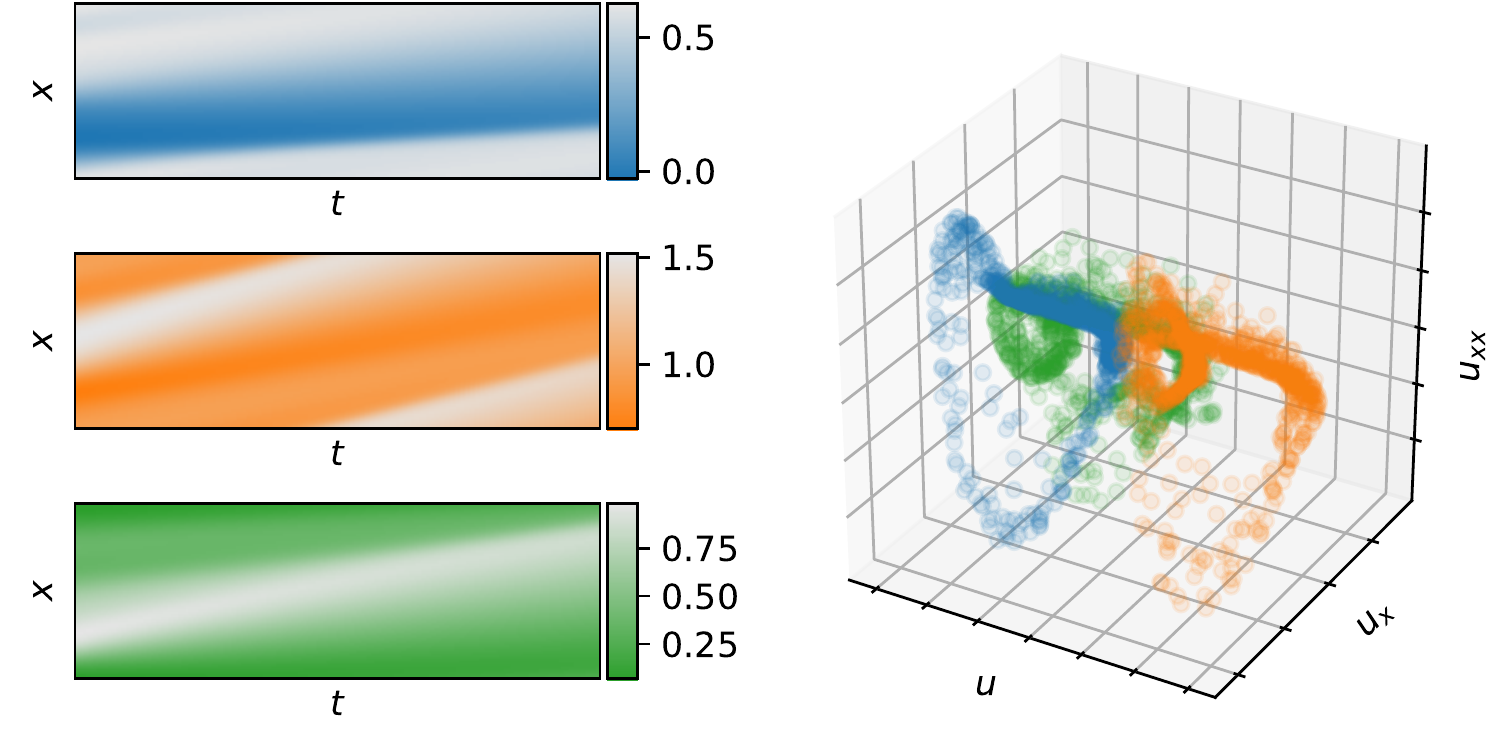}
         \caption{Experiment coverage.}
         \label{fig:coverage}
    \end{subfigure}
    \caption{
        (\subref{fig:BHPM}) Schematic of Bayesian hidden physics model.
        Leaves (green boxes) model spatiotemporal observables for distinct experiments.
        Automatic differentiation (blue boxes) is used to compute the required spatial derivatives of the leaves for the root.
        The root (brown box) models the unknown differential operator as a Gaussian process, trying to predict the $\pd{\hat{\vc u}}{t}$.
        (\subref{fig:coverage}) Three solutions (blue, orange, green) to Burgers' equation along with a scatter plot of their coverage in the input space of the root module, showing qualitatively where one can trust predictions based on the learned physics due to data coverage.
    }
    \label{fig:phpm}
\end{figure}

The key contribution of this work is to introduce a probabilistic approach to learning differential operators from data such that UQ for predictions based on the novel physics is possible.
Our approach, shown schematically in Fig.\ \ref{fig:BHPM} and explained in detail in Sec.\ \ref{sec:methodology}, combines information from multiple experiments to perform Bayesian inference directly on a nonparametric representation of the differential operator of interest.
Our approach also introduces a computationally tractable approach to propagating the uncertainty on the physics as represented above when solving novel initial-boundary value problems (IBVPs), inducing a distribution over solutions.
This UQ over predictions is a critical requirement for broader adoption of data-driven physics discovery.

The remainder of this paper is laid out as follows.
In Section \ref{sec:methodology}, we present the mathematical details of our approach.
In Section \ref{sec:related-work}, we review work related to our method.
In Section \ref{sec:examples}, we demonstrate our method on a number of example problems.
In Section \ref{sec:conclusion}, we summarize our findings, point out the limitations of our method, and identify promising avenues of future work.

\section{Methodology}
\label{sec:methodology}
In this section, we describe our approach to learning differential operators from a set of experiments.
We then show how the inferred operator can be used to solve the corresponding differential equation under arbitrary initial-boundary conditions.

\subsection{Problem statement}
We consider a dynamical system governed by a nonlinear partial differential equation
\begin{equation}
    \begin{aligned}
        \pd{u(\vc x, t)}{t} &= \scriptF [u(\vc x, t)] ~ \textrm{in} ~ \Omega_{st} = \Omega_s \times \Omega_t,
        \\
        u(\vc x, t=0) &= u_0(\vc x) ~ \textrm{on} ~ \Omega_s,
        \\
        \scriptB [u(\vc x, t)] &= b ~\textrm{on} ~ \partial \Omega_s \times \Omega_t,
    \end{aligned}
    \label{eqn:DE}
\end{equation}
where 
$\Omega_s \subset \reals^{d_s}$ is a $d_s$-dimensional spatial domain of interest, 
$\Omega_t \subseteq [0, \infty)$ is the temporal domain of interest,
$\scriptF$ is an unknown nonlinear differential operator, and
$\scriptB$ is some boundary operator.
We are provided as input datasets $\vc D^u = \{D_i^u\}_{i=1}^N$, each containing $n_{st}$ possibly-noisy measurements of $u$ in space and time, i.e.\ $D_i^u=\{x_j^{(i)}, t_j^{(i)}, \hat u_j^{(i)}\}_{j=1}^{n_{st}}$ where $\hat u_j^{(i)} \equiv \hat u(x_j^{(i)}, t_j^{(i)})$, each with possibly different $u_0$, $\Omega_{st}$, and $\scriptB$, i.e.\ particular solutions to distinct (IBVPs).
Our objective is to represent and infer the unknown $\scriptF$ such that one may solve new IBVP instances while quantifying the uncertainty of those predictions arising from our epistemic uncertainty of $\scriptF$ due to the finiteness of $\vc D^u$.

\subsection{Model definition}
Our model follows somewhat closely the Deep Hidden Physics Models approach of Raissi \cite{raissi2018deep}; the key differences in our model definition are that (1) we incorporate multiple datasets coming from distinct experiments within our model and (2) we seek to quantify and propagate the uncertainty associated with the learned differential operator through the Bayesian formalism.
We introduce our method for the case of physics that are represented by a first-order nonlinear partial differential equation in time with $d_s=1$, though the extension to multiple observables (a system of differential equations), higher temporal orders, and multiple spatial dimensions, is straightforward.
A schematic of our model, explained presently, is shown in Fig.\ \ref{fig:BHPM}.

The first component to our model is a set of ``leaf'' modules, each of which is a parametric function $u_i(x, t; \vc \theta_i^u), ~ i=1,\dots,N$.
Given parameters $\vc \theta_i^u$, the log-likelihood $L_i^u$ of the dataset $D_i^u$ is 
\begin{equation}
    L_i^u \equiv \log p(D_i^u|\vc \theta_i^u) = 
    \sum_{j=1}^{n_{st}} 
    \log p \left(
        \hat u_j^{(i)}|u(x_j^{(i)}, t_j^{(i)}; \vc \theta_i^u)
    \right).
    \label{eqn:leaf_ll}
\end{equation}
In this work, we use an i.i.d.\ Gaussian likelihood for each leaf with variances $\sigma_{u,i}^2,~i=1,\dots,N$.

The second component to our model is a ``root'' module that expresses the unknown nonlinear differential operator.
The key insight is to recognize that, under the assumption that $\scriptF$ is local and at most of order $K$ in $u(x, t)$ with respect to $x$, we can represent $\scriptF$ as an unknown (nonlinear) function $\pd{u}{t} = f(\vc v)$, where we define $\vc v = (u(x,t), \pd{u}{x}, \dots, \pd{^Ku}{x^K})$.
The domain of $f(\cdot)$ is $\Omega_f \subseteq \reals^{K+1}$.\footnote{
    For multiple spatial dimensions $d_s$, one must consider mixed derivatives.
    While the total number of combinations scales roughly as $\scriptO(d_s^k)$, this is not inherently problematic as typically $d_s \le 3$ in general and $k$ is usually not very large either. 
    The number of input dimensions to $f$ thus remains tractable in realistic settings.
}
We may then evaluate $f(\cdot)$ pointwise at values of $u(x, t)$, using automatic differentiation to obtain the required derivative elements of $\vc v$.
Throughout this work, we refer to both the operator over functions $\scriptF$ as well as its functional representation $f(\cdot)$ as the differential operator, using the mathematical symbol to disambiguate when necessary.

To formalize our uncertainty about $\scriptF$, we adopt a Bayesian probabilistic approach, modeling $f(\cdot)$ as a Gaussian process $\GP{\mu_f(\cdot; \vc \theta_{\mu_f})}{k_f(\cdot, \cdot; \vc \theta_{k_f})}$ with a Gaussian likelihood with variance $\sigma_f^2$.
In this work, we use a linear mean function and exponentiated quadratic kernel \cite{williams2006gaussian}.
Given a dataset 
$\vc D^f = \{ \mtx V \in \reals^{N n_{st} \times (K+1)}, \hat{\vc u}_t \in \reals^{N n_{st}} \}$ 
formed by computing the requisite derivatives to all leaves and concatenating, the marginal likelihood for the root module is
\begin{equation}
    L^f \equiv p(\hat{\vc u}_t | \mtx V) = \scriptN \left( \hat{\vc u}_t | \vc \mu(\mtx V), \mtx K_{ff} + \sigma_f^2 \mtx I_{Nn_{st} \times Nn_{st}} \right),
    \label{eqn:root_ll}
\end{equation}
where $\left(\mtx K_{ff}\right)_{ij} = k(\vc v_i, \vc v_j)$.
Eq.\ (\ref{eqn:root_ll}) is conditioned on $\vc \theta^u$ since $\vc D^f$ is generated by the leaves.

Reducing uncertainty about $\scriptF$ is thus analogous to posterior contraction on $f(\cdot)$.
Furthermore, we recognize that the credibility of the discovered differential equation may be quantified via the entropy $f(\cdot)$ at some $\vc v$ of interest; for GPs, this is intuitively connected to the data coverage in $\vc v$-space (see Fig.\ \ref{fig:coverage}).
Finally, the uncertainty on $f(\cdot)$ may be propagated when solving new IBVPs, thereby providing a formal means of quantifying the uncertainty associated with predictions (solving differential equations) with our learned physics as discussed in Sec.\ \ref{sec:methodology:predictions}.

The leaves and root are jointly trained by maximizing the joint log-likelihood of the model.
However, because of the large amount of data passing through the root, we opt to approximate $f(\cdot)$ using a sparse variational approximation following the familiar approach of \cite{hensman2013gaussian} with $n_u=128$ inducing inputs $\mtx V_u \in \reals^{n_u \times (K+1)}$ and corresponding inducing output variables $\vc f_u$.
Denoting as $p(\vc f)$ and $q(\vc f)$ the prior and variational posterior of the root GP at all of data points generated by the leaves, respectively, and evidence lower bound for the complete model is
\begin{equation}
    \log p \left(
        \left\{ \hat{\vc u}^{(i)}, \pd{\hat{\vc u}^{(i)}}{t} \right\}_{i=1}^N 
    \right)
    \ge
    \sum_{i=1}^N L_i^u
    +
    \doubleE_{q(\vc f)} \left[ \log p \left(\pd{\hat{\vc u}}{t} | \vc f) \right) \right] - KL\left(q(\vc f) || p(\vc f) \right).
    \label{eqn:elbo}
\end{equation}
Note that since the leaves are not treated probabilistically, only $L^f$ is variationally approximated,
though probabilistic leaves (e.g.\ Bayesian neural networks) could be used.
The model is trained by maximizing Eq.\ (\ref{eqn:elbo}) over the leaf and root parameters via gradient ascent, optionally with minibatching.

\subsection{Predictions and uncertainty quantification with the learned physics}
\label{sec:methodology:predictions}
After having trained the model described above, we can now use it to solve the discovered differential equations for arbitrary initial-boundary conditions.
Due to the stiffness associated with the problems we consider, we opt to use the method of physics-informed neural networks (PINNs) \cite{raissi2017physics1} as a black-box PDE solver.
The inputs to the PINNs solver are a deterministic differential operator $\tilde f$, and boundary operators $\tilde b_i(u, u_x, \dots),~i=1,\dots,n_B$.
The solution to the differential equation is parameterized by a neural network $\tilde u(x, t; \tilde{\vc \theta})$ whose parameters are optimized such that the squared residual of the domain and boundary operators are minimized over the domain and boundary of the problem, respectively. 
See the appendix for more details on the PINNs method and experiments quantifying the numerical accuracy of the method for the problems considered in this work.

Since the root module of our model represents a (variational) posterior over operators $q(f(\cdot))$ as a Gaussian process, one could sample $\tilde f(\cdot) \sim q(f(\cdot))$ using the root module and solving for each sample, we produce a distribution over solutions quantifying the effect of our epistemic uncertainty about $f(\cdot)$ on our predictions under the learned physics.

What remains is to determine how to obtain a sample of the operator function from the variational posterior $q(f(\cdot))$.
Because Gaussian processes are nonparametric, this cannot be done exactly in general.
Instead, we approximate a sample by introducing a set of conditioning points $\mtx V_c$ such that the \textit{augmented} posterior (conditioned on a sample of the inducing data as well as the conditioning points) is sufficiently narrow that its mean function may be regarded as a representative sample from $q(f(\cdot))$ \cite{oakley2002bayesian, bilionis2013multi}.
Formally, given a mean function $\mu(\cdot)$ and kernel function $k(\cdot, \cdot)$,
the Gaussian process posterior, conditioned on inputs $\mtx V_c \in \reals^{n_c \times (K+1)}$ and corresponding outputs $\vc f_c \in \reals^{n_c}$ is a Gaussian process
$f \sim \scriptG \scriptP (\mu_c(\cdot), k_c(\cdot, \cdot))$ with
\begin{equation}
    \mu_c (\vc v) = \mu(\vc v) + \vc k_{vc} \mtx K_{cc}^{-1} (\vc f_c - \mu(\mtx V_c)),
    k_c(\vc v, \vc v') = k(\vc v, \vc v') - \vc k_{vc} \mtx K_{cc}^{-1} \vc k_{vc}^\intercal,
    \label{eqn:gp_posterior}
\end{equation}
where 
$\left( \vc k_{vc} \right)_i = k(\vc v, \vc v_{c,i})$ 
and 
$\left( \mtx K_{cc} \right)_{i,j} = k(\vc v_{c,i}, \vc v_{c,j})$.
Crucially, if $k_c$ is small in some sense, then we may approximate $f(\cdot)$ by its (deterministic) mean function $\mu_c(\cdot)$.

The key question is thus how to pick the conditioning inputs $\mtx V_c$.
The following procedure provides one potential way to answer this question.
We first assign a uniform probability density over $\Omega_{st}$ and define $p(\vc v | \tilde{\vc \theta})$ as the pushforward density of $\scriptU[\Omega_{st}]$ under $\tilde u(\cdot; \tilde{\vc \theta})$.
Next, let
\begin{equation}
    L_c (\mtx V_c; \tilde{\vc \theta}) = \mathbb E_{p(\vc v | \tilde{\vc \theta})} \left[ \mathbb V \left[ p(f_{\vc v} | \mtx V_u, \mtx V_c, \tilde{\vc f}_u, \vc f_c) \right] \right],
    \label{eqn:condition_loss}
\end{equation}
where $p(f_{\vc v} | p(f_{\vc v} | \mtx V_u, \mtx V_c, \tilde{\vc f}_u, \vc f_c)$ is the marginal posterior of $f(\cdot)$ at $\vc v$, conditioned on $\mtx V_c$ as well as well as the inducing inputs $\mtx V_u$ and a sample from the variational posterior over the induced outputs $\tilde{\vc f}_u \sim q(\vc f_u)$.
Intuitively, $L_c$ quantifies the degree to which a function sampled from the current augmented posterior will differ from the augmented posterior mean on average, weighted to give extra importance to places in input space that are relevant to the solution to the current IBVP of interest.
The inner variance in Eq.\ (\ref{eqn:condition_loss}) can be computed in closed form, and the outer expectation may be quickly estimated via Monte Carlo.

Algorithm \ref{alg:ppinn} describes an iterative process by which one can successively refine a sample from $q(f(\cdot))$ as well as the corresponding solution $\tilde u(\cdot)$ subject to some given initial-boundary conditions.
The Algorithm alternates between refining $\tilde u(\cdot)$, given an operator, then refining the operator, given the current solution, until both converge.
By repeating this algorithm with different conditioning points, one can generate an ensemble of solutions that empirically describe the distribution of solutions to the given IBVP imparted by the uncertainty quantified by $q(f(\cdot))$.
As the tolerance $\delta_c$ is brought towards zero, $n_c$ will tend to increase, the final conditional $p(f(\cdot) | \mtx V_u, \vc f_u, \mtx V_c, \vc f_c)$ will be narrower, and the augmented posterior mean will be a more faithful sample to the true posterior over functions.

\begin{algorithm}[htb]
    \textbf{Require}:
    Initial condition $u_0(\cdot)$,
    boundary conditions $\{\tilde b_i(\cdot) = 0\}_{i=1}^{n_B}$, 
    GP mean $\mu(\cdot)$ and kernel $k(\cdot, \cdot)$,
    initial conditioning points $\{\mtx V_{c,0}, \vc f_{c,0}\}$,
    number of points to add per iteration $n_{c,new}$,
    tolerance $\delta_c$.
    
    \textbf{Ensure}: 
    Approximate operator sample $\tilde f(\cdot) \sim q(f(\cdot))$ and
    optimized solution network parameters $\tilde{\vc \theta}$.
    \begin{algorithmic}
        \STATE
        $\mtx V_c \leftarrow \mtx V_{c,0}$,
        $\vc f_c \leftarrow \vc f_{c,0}$.
        \LOOP
            \STATE $\tilde f(\cdot) \leftarrow \mu_c(\cdot)$ using Eq.\ (\ref{eqn:gp_posterior}).
            \STATE $\tilde{\vc \theta} \leftarrow \texttt{PINN} \left( \tilde f(\cdot), \{b_i(\cdot)\}, u_0(\cdot) \right)$.
            \STATE Compute $L_c (\mtx V_c; \tilde{\vc \theta})$ using Eq.\ (\ref{eqn:condition_loss}).
            \IF{$L_c < \delta_c$}
                \STATE \textbf{break}
            \ENDIF
            \STATE Update $\mu \leftarrow \mu_c$, $k \leftarrow k_c$.
            \STATE Pick a new set of conditioning points 
            $\mtx V_{c,new} = \argmin_{\reals^{n_{c,new} \times (K+1)}} L_c \left( (\mtx V_c^\intercal, \mtx V_{c,new}^\intercal)^\intercal; \tilde{\vc \theta} \right)$ via gradient descent on $L_c(\mtx V_c)$.
            \STATE Sample $\vc f_{c,new}(\mtx V_{c,new})$ using Eq.\ (\ref{eqn:gp_posterior}).
            \STATE $\mtx V_c \leftarrow (\mtx V_c^\intercal, \mtx V_{c,new}^\intercal)^\intercal$,
            $\vc f_c \leftarrow (\vc f_c, \vc f_{c,new})$.
        \ENDLOOP
        \RETURN $\tilde f(\cdot), \tilde{\vc \theta}$.
    \end{algorithmic}
 \caption{Sample a solution to an IBVP under a random differential operator.}
 \label{alg:ppinn}
\end{algorithm}

\section{Related work}
\label{sec:related-work}
Our work is closely related to that of Raissi \cite{raissi2018deep}, who first proposed that unknown differential operators might be represented as functions learned from data. 
However, this prior work lacked a formal approach to understanding the credibility of the learned operator and could only assess it through trial and error, requiring access to ground truth solves for comparison.
Furthermore, \cite{raissi2018deep} is unable to answer definitively why certain predictions fail, merely offering a qualitative explanation about the ``richness'' of a dataset.
Our work provides a formal approach to posing and (as shown below) answering these questions.

Earlier work by Rudy et al.\ \cite{rudy2017data} propose to learn differential equations from data, using sparse regression on a pre-defined library of terms.
However, they use symbolic differentiation on local polynomial fits to obtain gradient information; this crude procedure makes it challenging in practice to learn from noisy data. 
By contrast, our approach allows uses to use highly expressive function models while equipping the leaves with likelihoods to separate inference regarding the underlying function from the noise corrupting the observations.
The benefit of having a general compatibility with differentiable models for each experiment was also leveraged by Atkinson et al.\ \cite{atkinson2019data}.
Furthermore, \cite{rudy2017data} mention that their approach can be applied to subsampled data, but they do not discuss the implications of how the data are subsampled, whereas our formal quantification of epistemic uncertainty now allows one to consider this question rigorously.

There have been some works applying UQ to certain aspects of physics-informed machine learning.
Raissi et al.\ \cite{raissi2018numerical} consider propagating uncertainty in initial conditions, Yang et.\ al \cite{yang2020b} explore use Bayesian neural networks to quantify epistemic uncertainty in the function model, and Atkinson et al.\ \cite{atkinson2019data} conduct Bayesian inference on the parameters of proposed symbolic differential operators.
Our work is different from the former two in that the target of our uncertainty quantification is the physics itself by way of the nonparametric distribution over functions representing the unknown differential operator.
We differ from the latter in that we show how to propagate the quantified uncertainty when solving novel IBVPs.
Also, it is unclear whether Bayesian inference over parameters in symbolic physics necessarily confers appropriate semantics regarding uncertainty in operator space for credible uncertainty propagation.
By using a Gaussian process as a nonparametric distribution over operators, we ensure that its posterior possesses desirable semantics.

Our use of a Gaussian process to represent the differential operator is somewhat similar to Gaussian process state-space models \cite{wang2006gaussian, frigola2013bayesian, frigola2014variational, eleftheriadis2017identification, ialongo2018closed, ialongo2018non}, though our work is different in that instead of a finite-dimensional state in discrete time we model the evolution of a spatial field in continuous time.
The insight to model the unknown differential operator as a function acting pointwise on the solution function and its derivatives is not obvious, and while previous work on tractable coherent multi-step predictions is similar in spirit to our goal of sampling solutions to novel IBVPs, it is not immediately clear how existing methods can be made to accomplish this in our setting.

\section{Examples}
\label{sec:examples}
In this section, we demonstrate our method on several types of examples to illustrate how knowledge and uncertainty associated with discovered physics may be quantified.
Code to reproduce our results will be made available on GitHub upon publication.
Details involving the setup of our experiments including network architectures and training schedules can be found in the Appendix.
In all cases, ground truth data are generated with a Python port of the Chebfun package \cite{driscoll2014chebfun}; details are included in the Appendix.
We consider two nonlinear partial differential equations: Burgers' equation, $u_t = \scriptF[u] = -u u_x + 0.1 u_{xx}$; and the Korteweg-de Vries (K-dV) equation, $u_t = \scriptF[u] = -u u_x - u_{xxx}$.

\begin{figure}[hbt]
    \centering
    \includegraphics[width=0.25\textwidth]{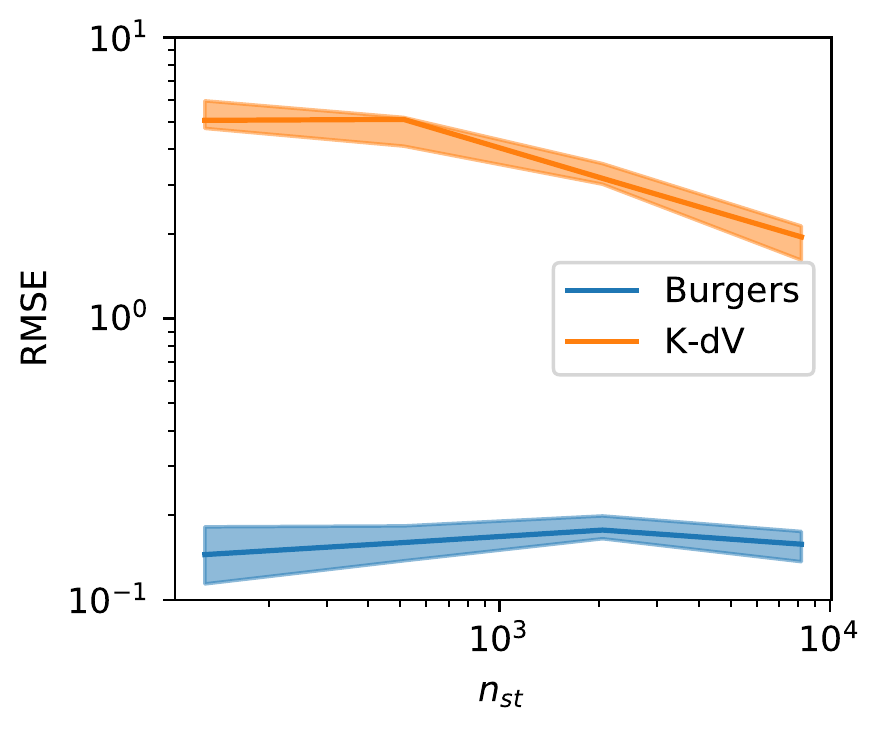}
    \includegraphics[width=0.25\textwidth]{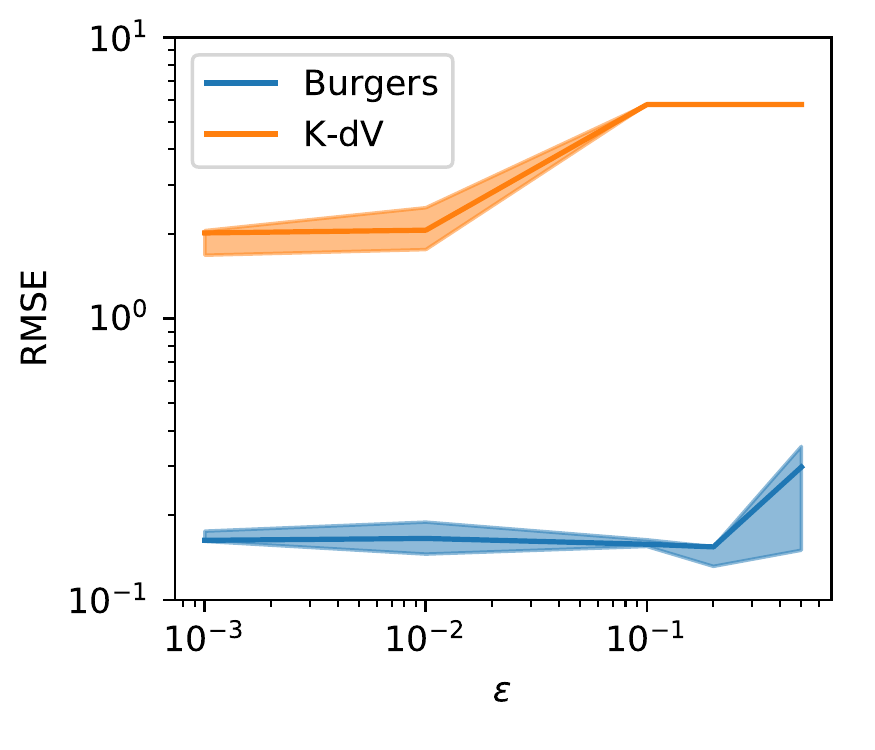}
    \includegraphics[width=0.25\textwidth]{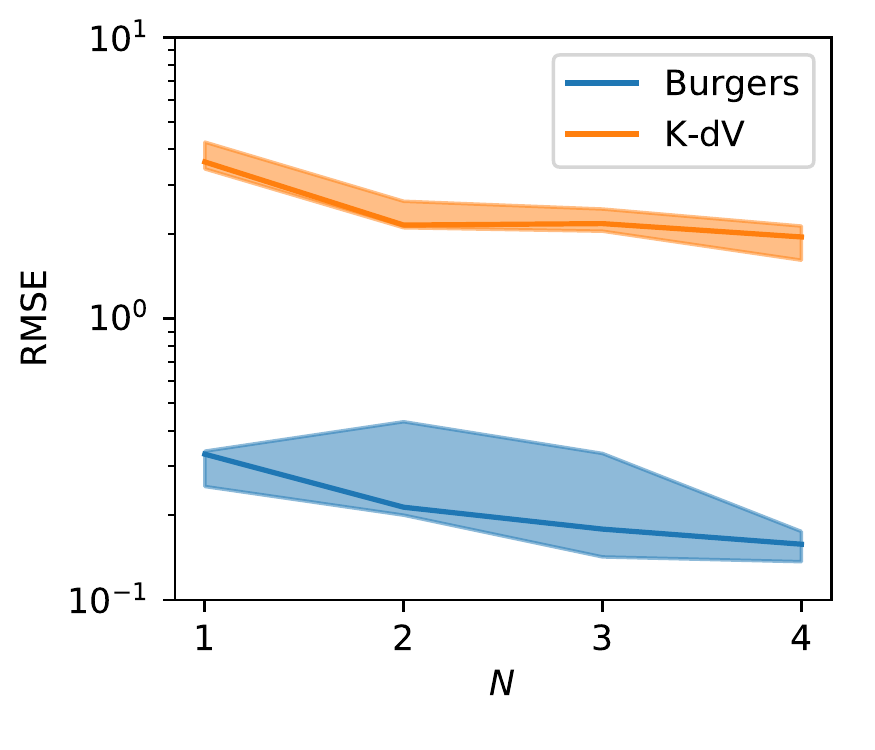}
    \\
    \includegraphics[width=0.25\textwidth]{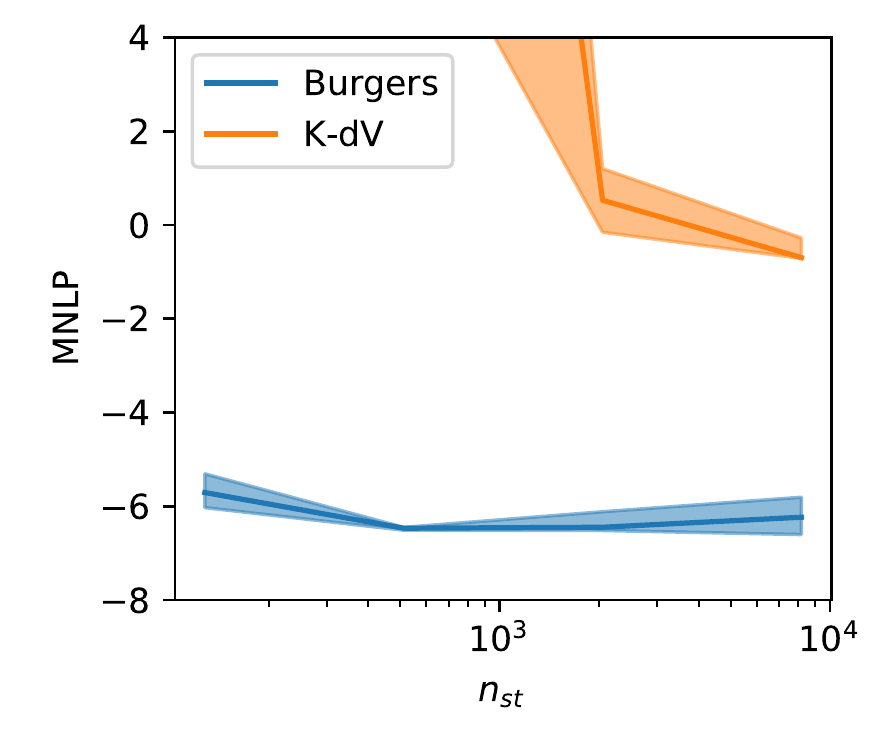}
    \includegraphics[width=0.25\textwidth]{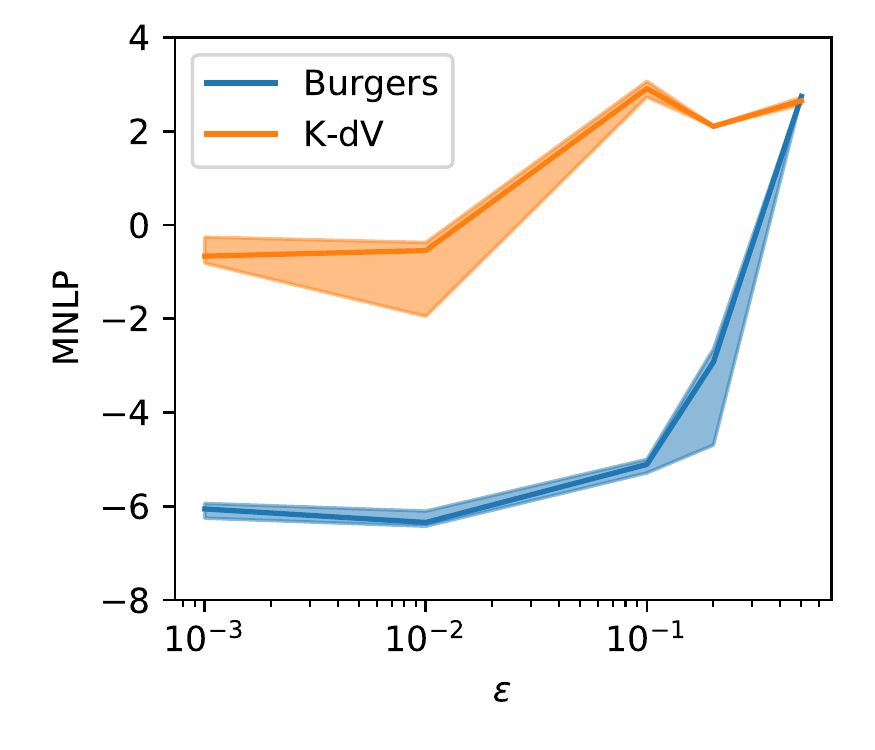}
    \includegraphics[width=0.25\textwidth]{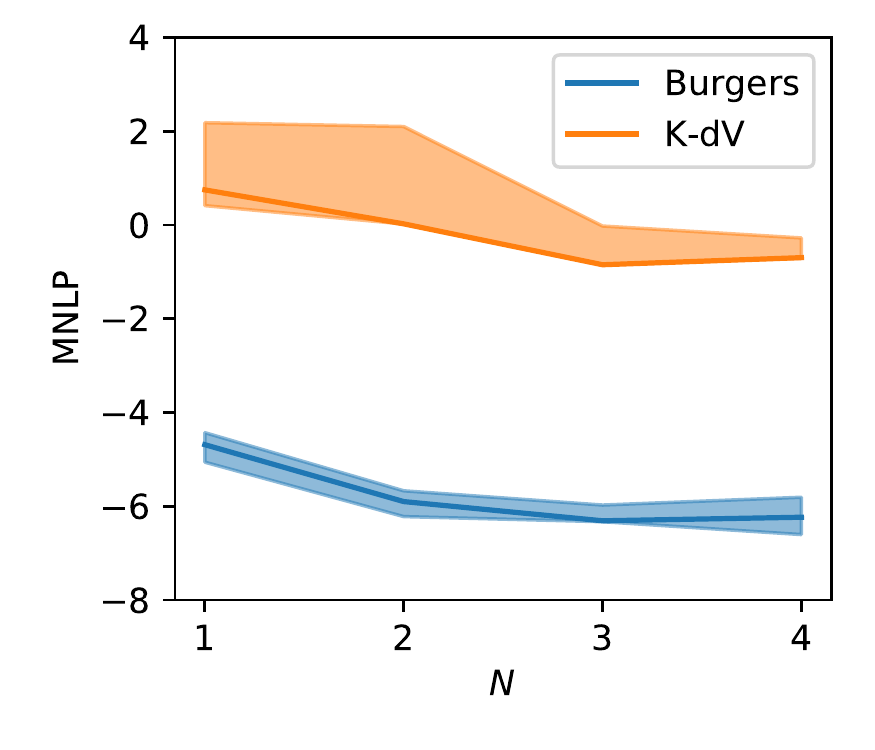}
    \caption{
        Predictive accuracy of the trained roots in terms of RMSE (top) and MNLP (bottom) as the number of measurements per experiment (left), noisiness of the measurements (center), and number of experiments (right) are varied.
        MNLP for K-dV becomes very large when $n_{st}$ is small.
    }
    \label{fig:examples:root_accuracy}
\end{figure}

\textbf{Quantifying the accuracy of the learned physics}
Because our approach represents $\scriptF$ as an unknown function $f(\cdot)$, we may test the accuracy of our inference through the framework of standard supervised learning, using $(\vc v, \hat{u_t})$ pairs from held-out experiments as test data.
The BHPM model is trained on a set of $N$ experiments with $n_{st}$ measurements per experiment, randomly subsampled from the $n_s n_t$ points computed by the solver and corrupted with Gaussian noise with standard deviation $\epsilon$.
A held-out set of $N^*=10$ experiments is used to generate inputs and outputs for testing the operator learned by the root.
Spatial derivatives are computed by application of symbolic differentiation to the Fourier series used by our spectral element method, and the target time derivative is computed by application of the ground-truth operator.
We quantify performance using the root mean squared error (with respect to the posterior mean of $f(\cdot)$) and median negative log probability of the predictive posterior.\footnote{We opt for the median rather than the mean because we found that the results tend to be otherwise dominated by a small number of outliers.}
We first aim to demonstrate that the learned physics trends in the expected ways as we increase $n_{st}$ (improves), $\epsilon$ (degrades), and $N$ (improves).
For each sequence of experiments, the parameters not being varied were set to $N=4$, $n_{st}=8192$, and $\epsilon = 0$.
Figure \ref{fig:examples:root_accuracy} collects our results, repeating each experiment three times.
While the expected trends can be observed, they are rather mild in most cases.
We hypothesize that an adaptive experimental design driven by our approach's quantification of the uncertainty in the learned physics could substantially increase the performance of our model over the random acquisition strategy we employ here.

\textbf{Convergence of solution samples due to operator sample refinement}
Turning our attention to making predictions with the learned physics, we now empirically investigate the convergence of the solution function $\tilde u(\cdot)$ as the conditioning set $\{ \mtx V_c, \vc f_c\}$ is grown using Algorithm \ref{alg:ppinn}.
We consider the case of a synthetic distribution over operators with signature $f(u, u_x, u_{xx})$ given by a Gaussian process with prior mean function and kernel 
$\mu_f(u, u_x, u_{xx}) = 0.1 u_{xx}$, 
$k(\vc v, \vc v') = 0.1^2 \exp \left[ \sum_{i=1}^{K+1} (v_i - v_i')^2 \right]$, 
conditioned by a single initial deterministic observation $\mtx V_{c,0} = [0,0,0]$, $f_{c,0} = 0$.
Thus, the initial GP posterior mean corresponds to the 1D heat equation.
We solve on $\Omega_s = [-\pi, \pi]$, $\Omega_t=[0, 10]$, using $u_0(x) = \sin(x)$ and periodic BCs in space.
We add $n_{c,new}=8$ conditioning points per iteration.
Intuitively, decreasing $\delta_c$ will increase $n_c$ and the variability in the distribution over solutions, whereas higher tolerances will tend to underestimate the true uncertainty.

\begin{figure}[hbt]
    \centering
    \begin{subfigure}[b]{0.6\textwidth}
        \centering
        \includegraphics[width=0.32\textwidth,clip]{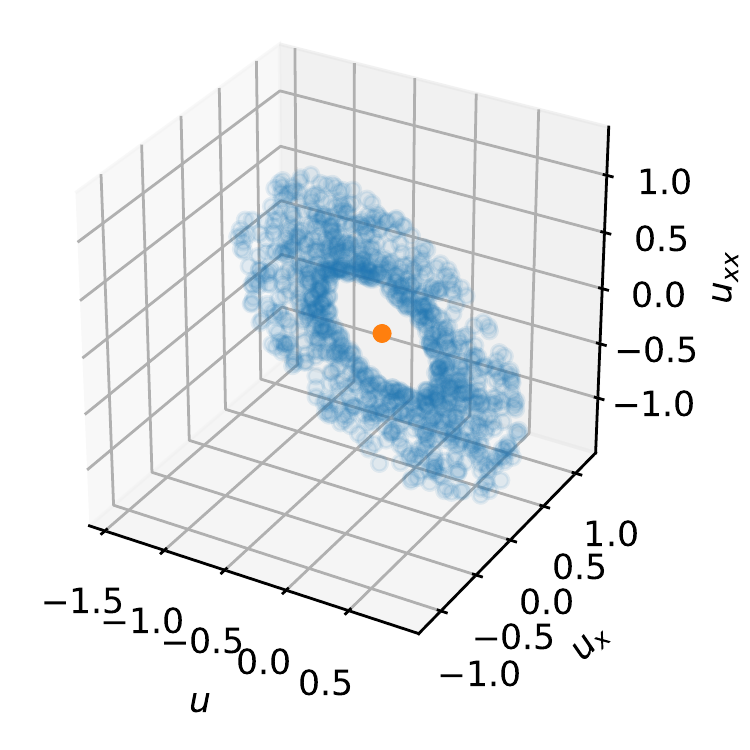}
        \includegraphics[width=0.32\textwidth,clip]{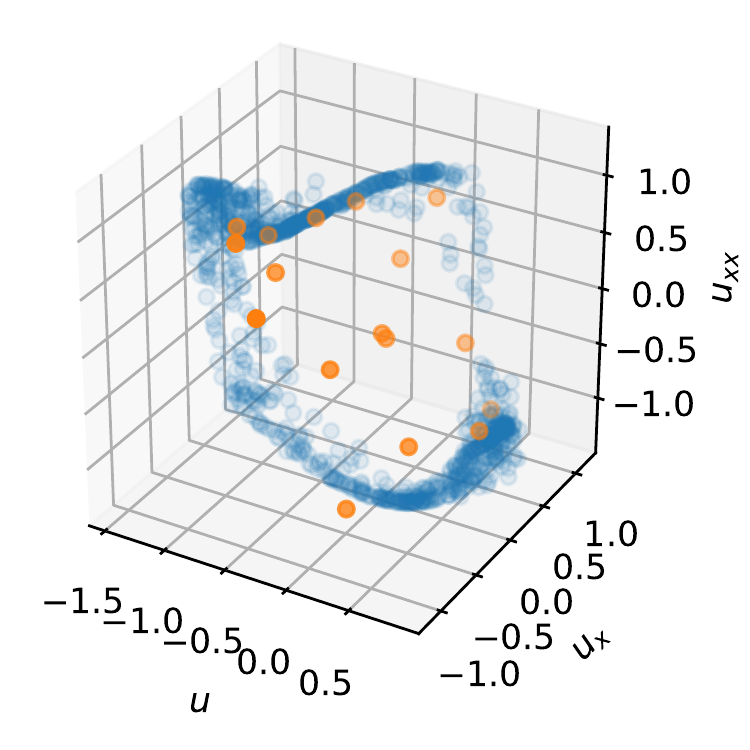}
        \includegraphics[width=0.32\textwidth,clip]{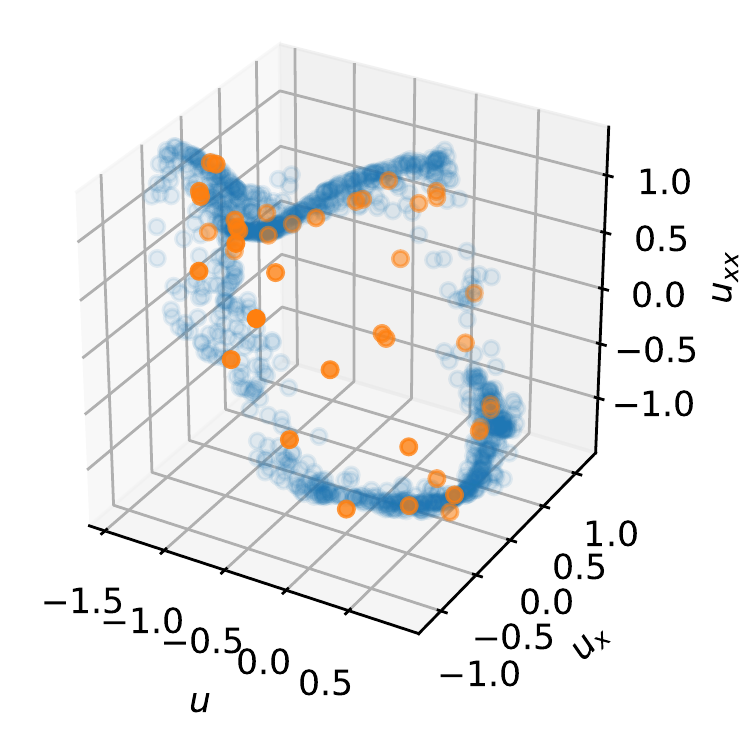}
        \\
        \includegraphics[width=0.32\textwidth,clip]{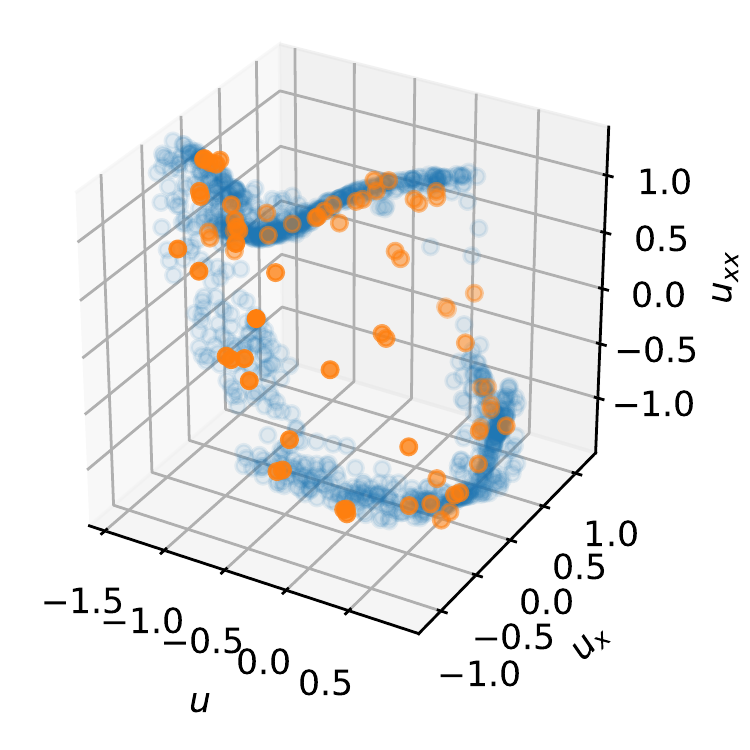}
        \includegraphics[width=0.32\textwidth,clip]{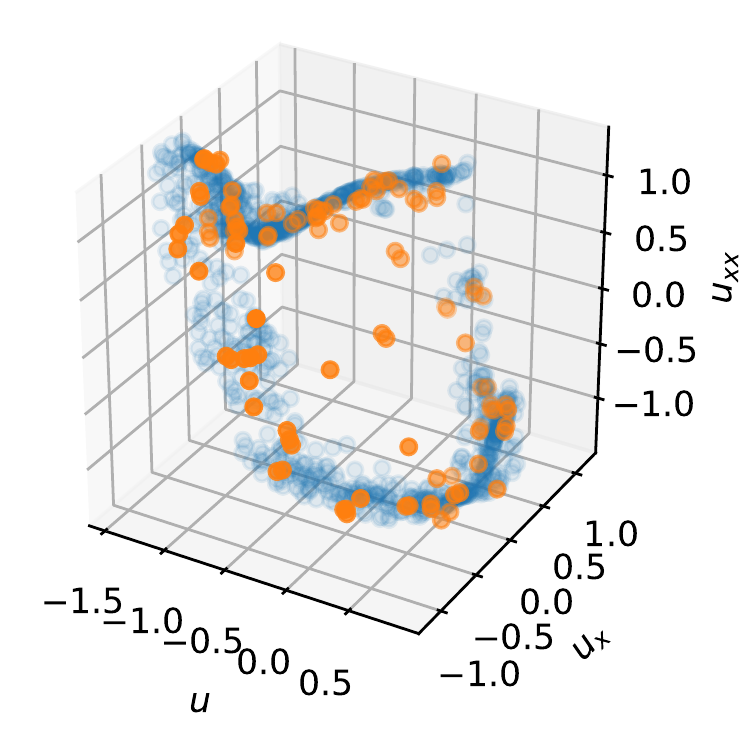}
        \caption{Convergence in $\vc v$-space.}
        \label{fig:refine:state}
    \end{subfigure}
    \begin{subfigure}[b]{0.35\textwidth}
        \centering
        \includegraphics[width=\textwidth]{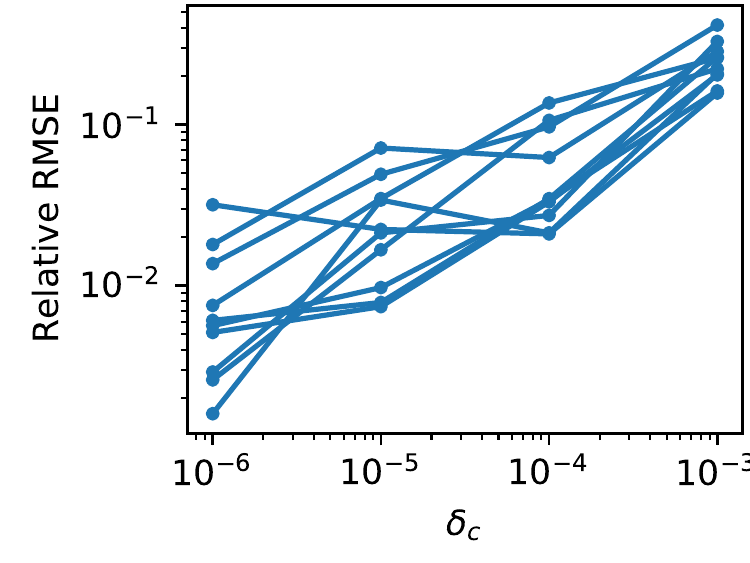}
        \caption{Convergence of $\tilde u(\cdot)$.}
        \label{fig:refine:tol}
    \end{subfigure}
    \caption{
        (\subref{fig:refine:state}) Convergence in $\vc v$-space of a solution sample as a function of tolerance.
        From left to right, top to bottom: $\delta_c=10^{-2}$, $10^{-3}$, $10^{-4}$, $10^{-5}$, and $10^{-6}$.
        Samples of $\vc v(x,t)$ with $(x,t) \sim \scriptU[\Omega_{st}]$ (blue) and selected conditioning inputs $\mtx V_c$ (orange) are shown.
        (\subref{fig:refine:tol}) Relative root mean squared error between successive solutions as the tolerance is decreased.
        Each line connects successive refinements of one sample from $q(f(\cdot))$.
    }
    \label{fig:gpinn:state_scatter}
\end{figure}

Figure \ref{fig:refine:state} shows the convergence of a single solution sample in $\vc v$-space as the tolerance $\delta_c$ is decreased, and Fig.\ \ref{fig:refine:tol} quantifies the relative RMSE between a solve obtained by running Algorithm \ref{alg:ppinn} to a tolerance $\delta_c$ versus the result obtained by termination at $10 \delta_c$, measured on a uniform grid of $128 \times 128$ points covering $\Omega_{st}$.

\begin{figure}[hbt]
    \centering
    \begin{subfigure}[b]{0.9\textwidth}
        \centering
        \includegraphics[width=\textwidth]{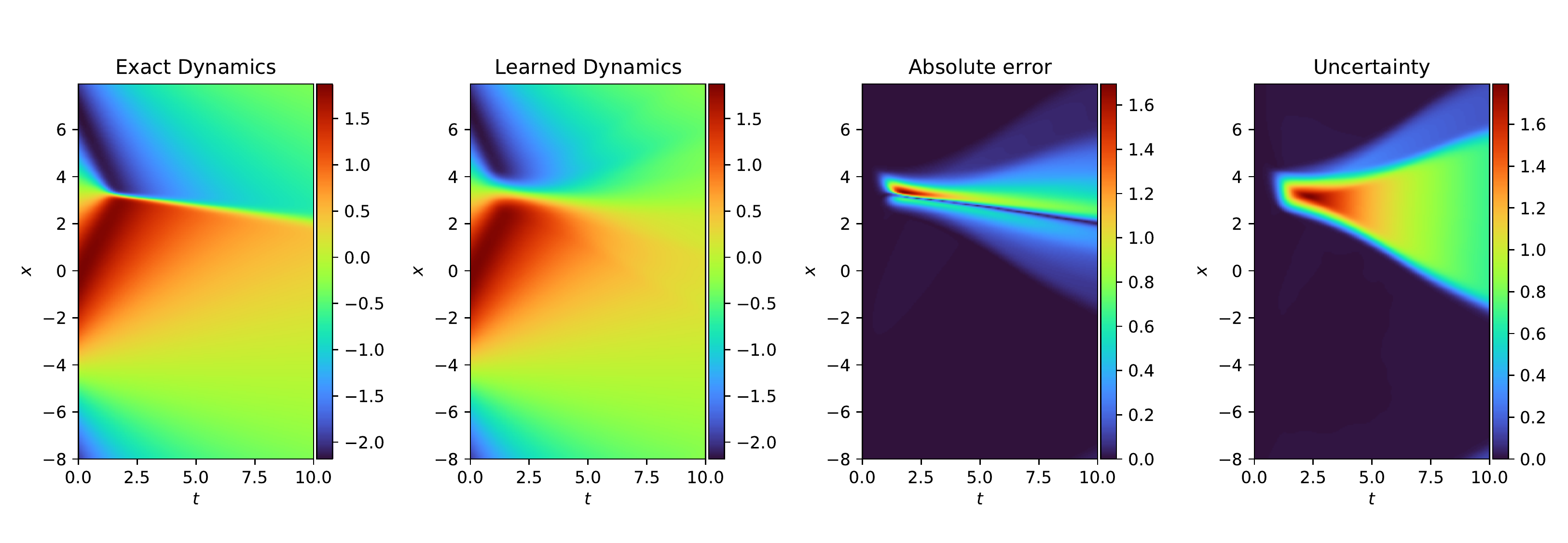}
        \vspace{-0.25in}
        \\
        \includegraphics[width=\textwidth]{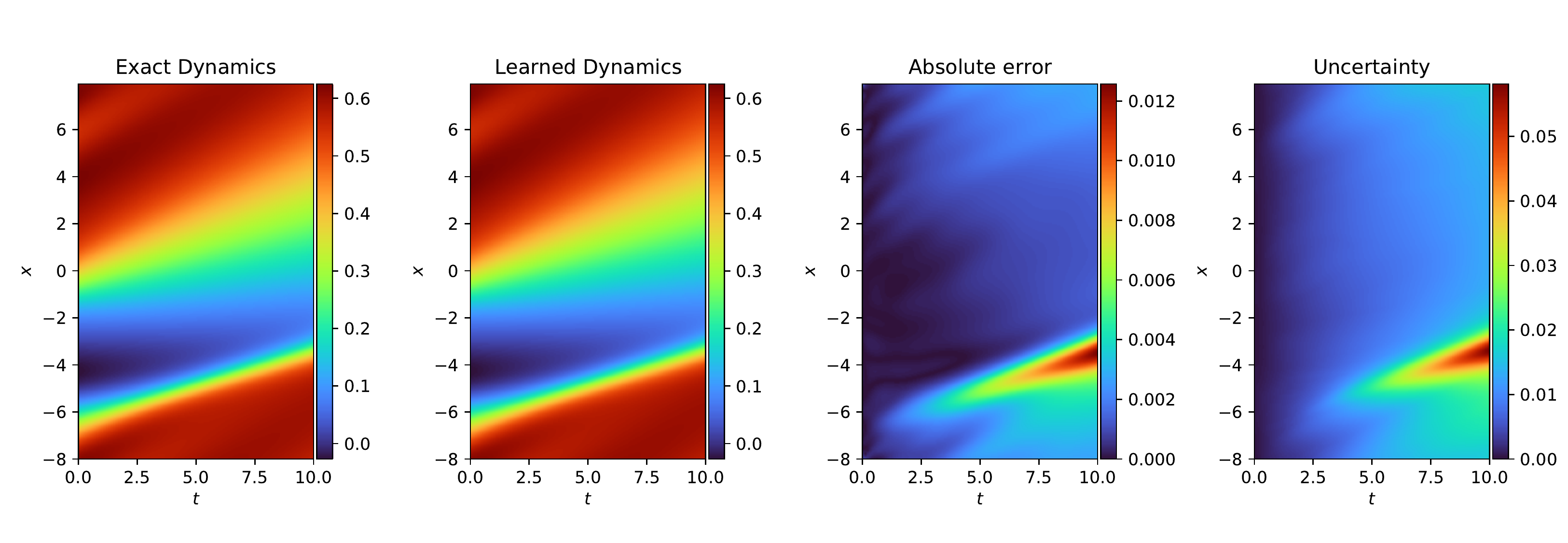}
        \vspace{-0.25in}
        \caption{Burgers' equation.}
        \label{fig:burgers_hi}
    \end{subfigure}
    \\
    \begin{subfigure}[b]{0.9\textwidth}
        \centering
        \includegraphics[width=\textwidth]{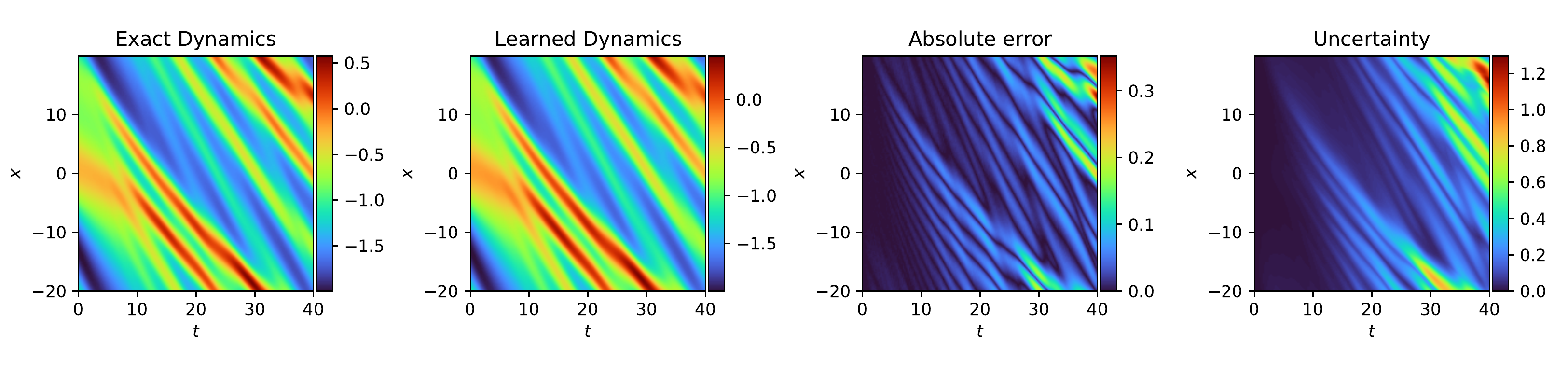}
        \vspace{-0.25in}
        \\
        \includegraphics[width=\textwidth]{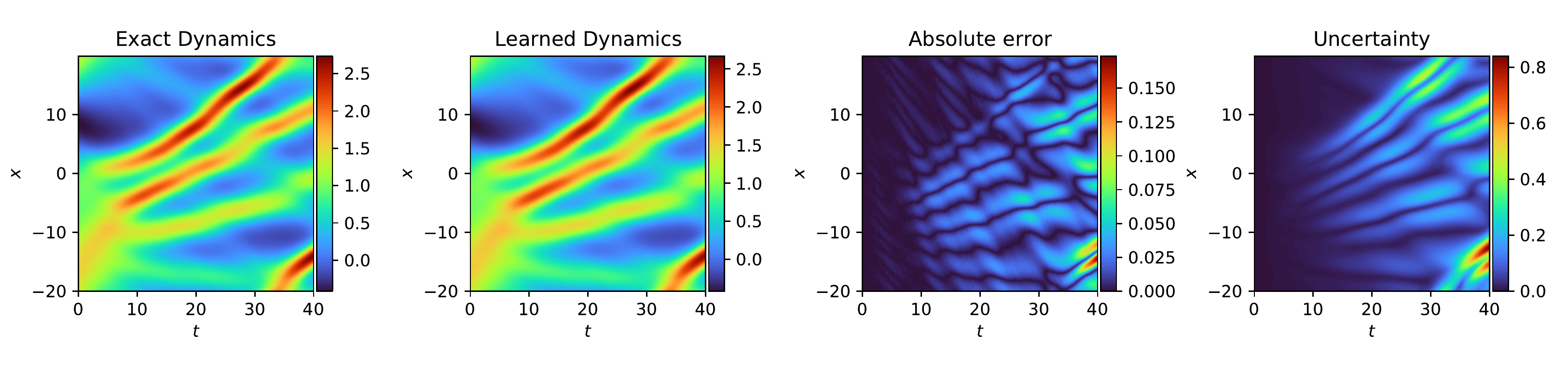}
        \vspace{-0.25in}
        \caption{K-dV equation.}
        \label{fig:kdv_hi}
    \end{subfigure}
    \caption{Predictions using the learned physics under new initial conditions.}
    \label{fig:up}
\end{figure}

\textbf{Propagating uncertainty to solutions to novel IBVPs}
We now demonstrate the approach described in Sec.\ \ref{sec:methodology:predictions} using the operator posteriors learned from data from Burgers' and the Korteweg-de Vries equations.
Physics were learned from $N=4$ solutions with $n_{st}=4096$ measurements randomly drawn from each example.
Figure \ref{fig:up} shows predictions on new initial conditions; periodic boundary conditions are used as before.
Predictive means and uncertainties corresponding to half of the empirical 95\% confidence interval are shown based on $32$ samples of the operator posterior.
We see that by quantifying and propagating uncertainty on the physics itself, we inherit nuanced and well-calibrated UQ on the solutions.
In these experiments, we find that $L_c$ tends to be on the order of floating-point precision when conditioning only on $\mtx V_u$ and the corresponding samples from $q(\vc f_u)$.
This makes sense given that the variational approach \cite{titsias2009variational} implies that inducing points act as sufficient statistics of the true posterior.
If one were to make predictions on out-of-distribution initial conditions, then the inducing points may not cover the correct region of $\vc v$-space and additional condition points may be required.

\section{Conclusions and Discussion}
\label{sec:conclusion}
We have demonstrated a method to infer nonlinear partial differential equations from data while quantifying the epistemic uncertainty due to the finite nature of the experimental data.
We further saw that this uncertainty may be propagated when making predictions with the learned physics.

Our work illuminates a number of fruitful extensions.
First, while we used random sampling methods to collect training data, our formalism paves the way for using information-theoretic techniques for experimental design in the context of nonparametric inference over differential operators.
Second, while our method provides a nonparameteric description of the learned physics, one might distill a symbolic representation of $f(\cdot)$ by applying existing techniques \cite{rudy2017data, atkinson2019data, cranmer2019learning} to the operator discovered by our method.
Furthermore, the posterior uncertainty on $f(\cdot)$ might remain useful to recognize when predictions made with the symbolic distillation of the physics are not reliable.

We only considered in this work first-order dynamics where all observables entering into $\scriptF$.
It is straightforward to adapt our method to discovering higher-order dynamics e.g.\ $u_{tt} = \scriptF[u]$.
Additionally, it is known \cite{takens1981detecting, somacal2020uncovering} that hidden variables may be indirectly by using multiple time lags of the observables; this is used commonly when learning black-box models of dynamics from snapshots, e.g.\ when using pairs of consecutive images in time to indirectly represent velocities.
Additionally, previous work \cite{raissi2018deep} has shown that the absence of required input terms may be readily diagnosed empirically.

While a Gaussian process are typically regarded as having good inductive biases for representing uncertainty in functions, we suspect that there is considerable room for improvement over the Gaussian kernel and linear mean function used in this work for the root module.
Whereas kernel functions are usually given more attention due to their effect on the local properties of the functions (e.g.\ smoothness and length scales), it is important capture the limiting behavior correctly, implying that the choice of mean function may be quite impactful.

A key counterpart to the focus of this work on discovering new physics is the development of techniques for solving novel PDEs.
Solving PDEs with black-box operators remains in its infancy though advances are being made to better understand the challenges of PINN-based algorithms \cite{wang2020understanding} as well as empirical findings improving performance \cite{jagtap2020adaptive,shukla2020physics,jin2020nsfnets}.

\section*{Broader Impact}

We anticipate in general that research in scientific machine learning, including work to use ML to discover new physics may become a defining tool of the modern physical sciences.
Given its wide potential application, we believe it is reasonable to expect both significant positive and negative consequences, as has been the legacy of the physical sciences.
Rather than speculating broadly about the future legacy of the physical sciences, accelerated by the targeted application of ML, we would like to instead address the impact of ML as a tool of automation on the role of the modern scientist as well as the impact of uncertainty quantification in the operationalization of novel physics in technology.

Regarding the first, we do not believe that the potential to apply ML to automate aspects of the scientific process poses by itself an existential threat to the job of a scientist.
By its definition, science seeks to understand what is novel, and so it seems reasonable that there will be a continuing need for the scientist to interact with novel experimental settings as well as frame the questions that may be answered using approaches such as ours.
Moreover, the automation of some technical analysis may accentuate the role of the scientist in making value judgments regarding various lines of inquiry.

Regarding the second, we believe that the ability to reason quantitatively and formally about the credibility and applicability of novel physics is fundamental to its safe integration into larger engineering technologies and systems.
Without this, it is hard to believe that critical applications involving novel physics could be prudent without considerable empirical experimentation and testing with the associated costs in time and money.
Simultaneously, we expect that uncertainty quantification in the context of novel physics could be an important safeguard in its application, enabling engineers to be conscious stewards of their technology and prevent unintended misuse e.g.\ by pushing operational conditions beyond those in which the underlying physics have been verified.

\begin{ack}

This material is based upon work supported by the Defense Advanced Research Projects Agency (DARPA) under Agreement No.\ HR00111990032.
\end{ack}

\newpage
\appendix
\section{Datasets}
\begin{figure}[hbt]
    \centering
    \begin{subfigure}[b]{\textwidth}
         \centering
         \includegraphics[width=0.19\textwidth,clip]{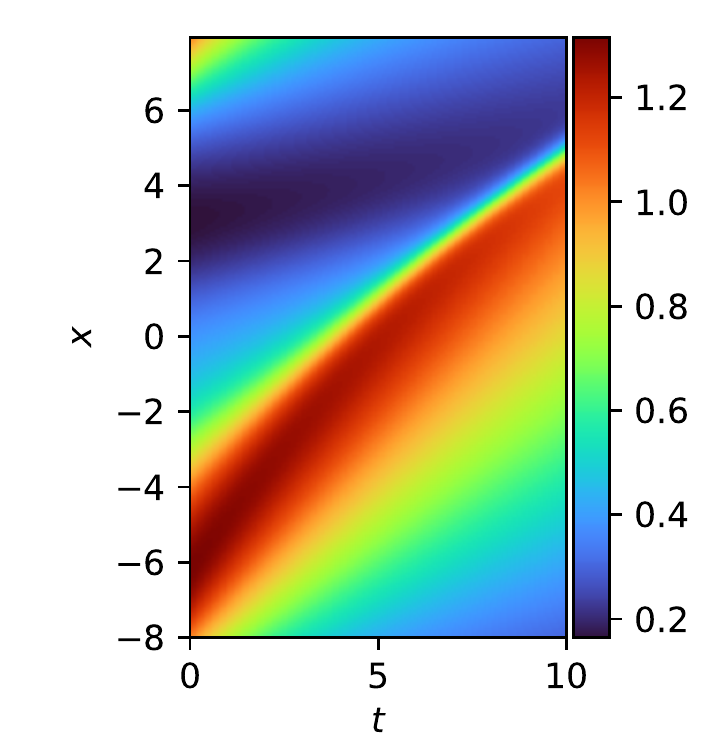}
         \includegraphics[width=0.19\textwidth,clip]{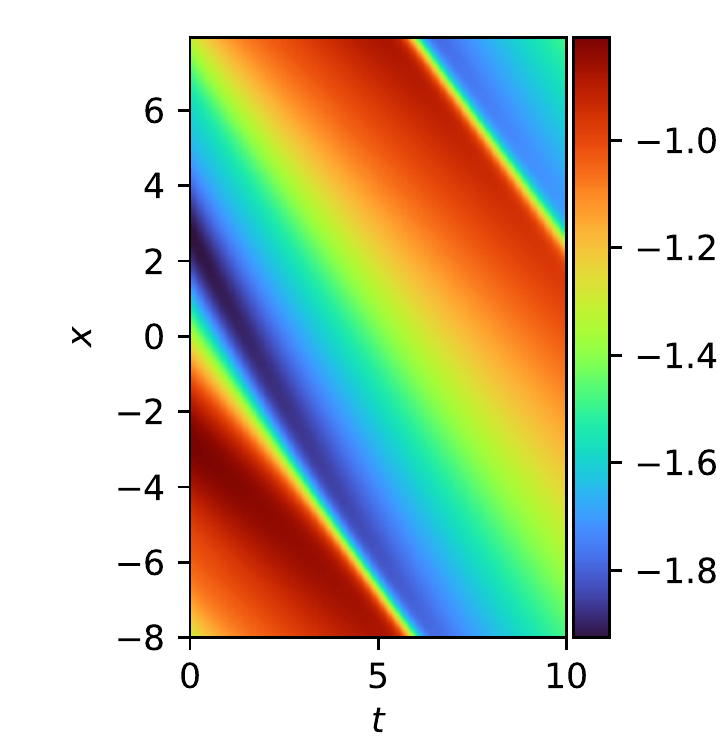}
         \includegraphics[width=0.19\textwidth,clip]{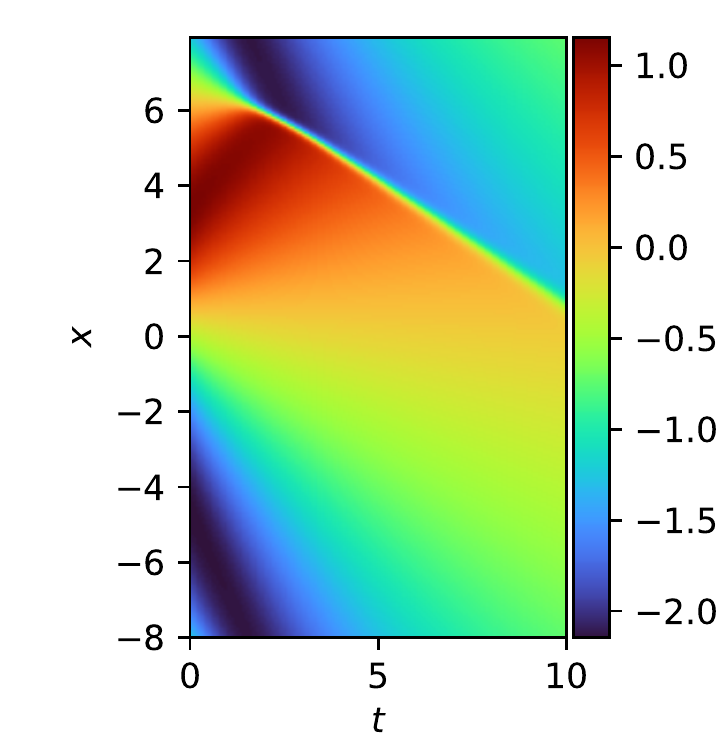}
         \includegraphics[width=0.19\textwidth,clip]{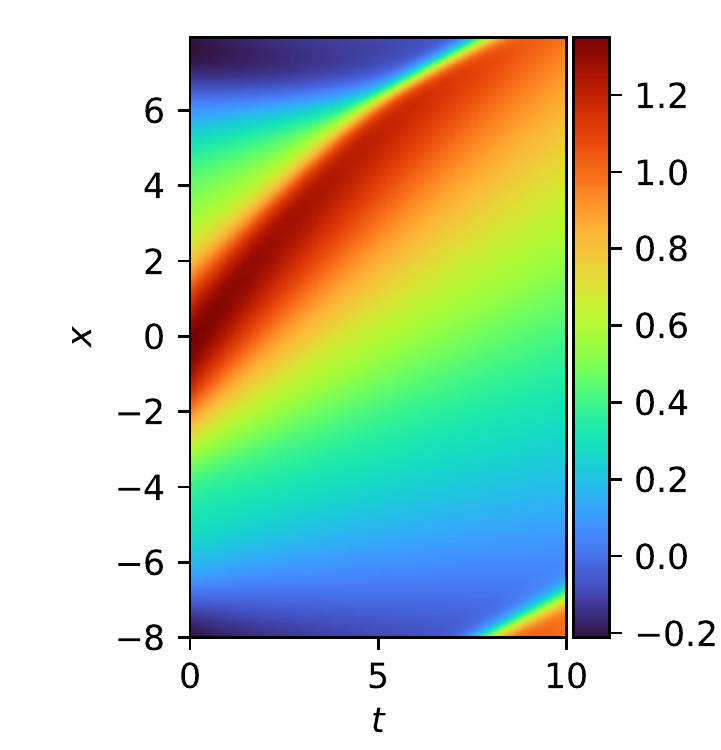}
         \includegraphics[width=0.19\textwidth,clip]{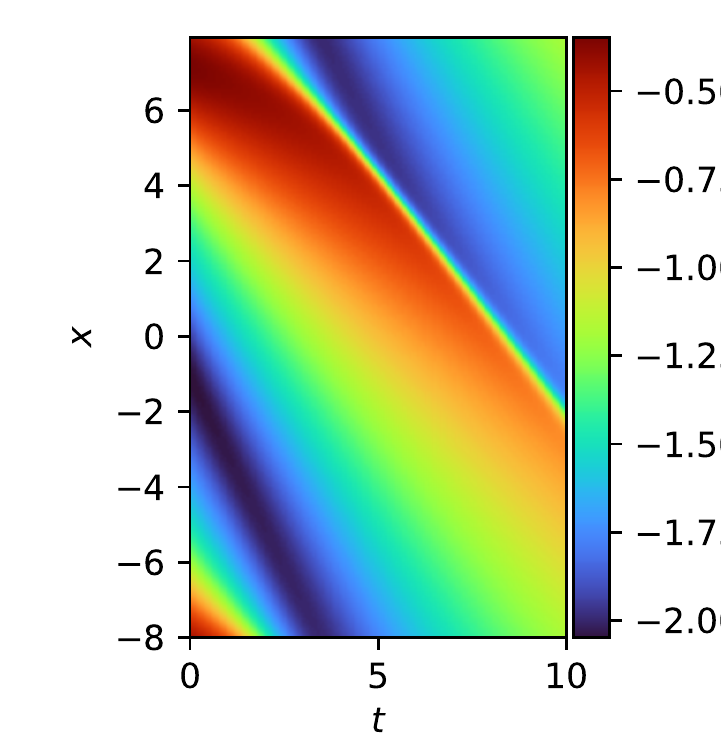}
         \includegraphics[width=0.19\textwidth,clip]{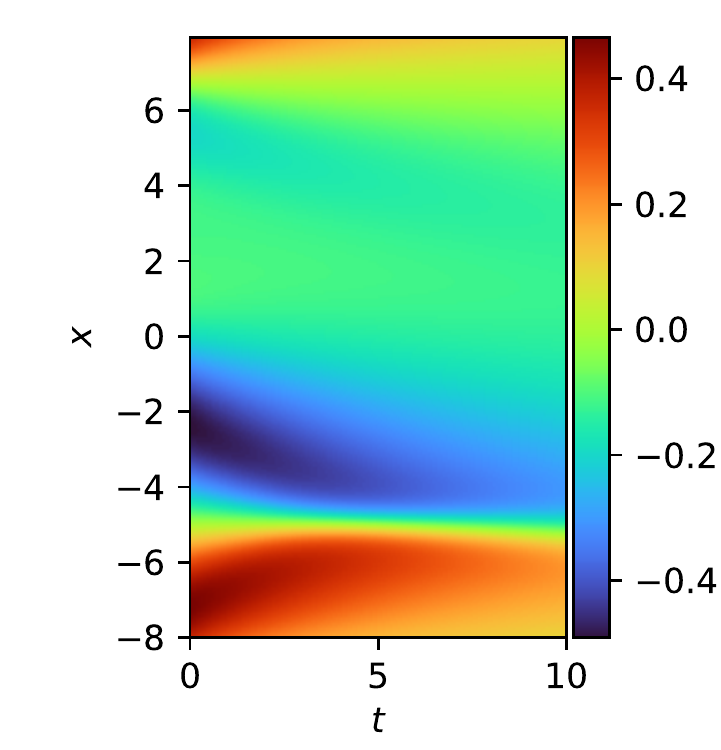}
         \includegraphics[width=0.19\textwidth,clip]{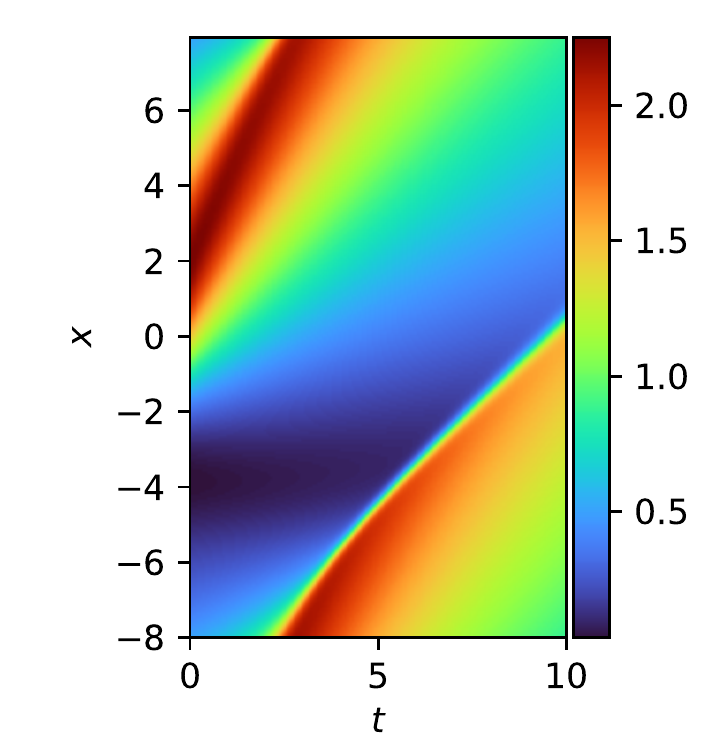}
         \includegraphics[width=0.19\textwidth,clip]{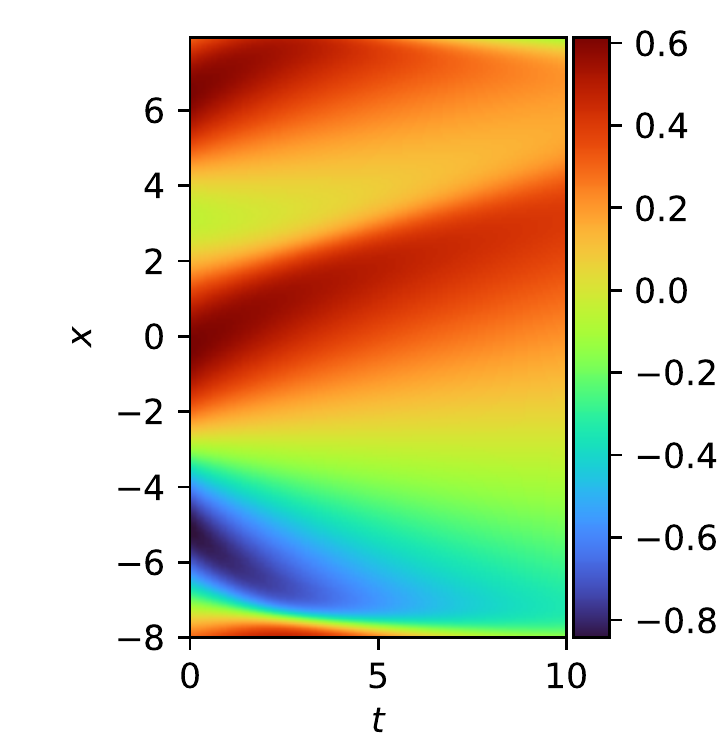}
         \includegraphics[width=0.19\textwidth,clip]{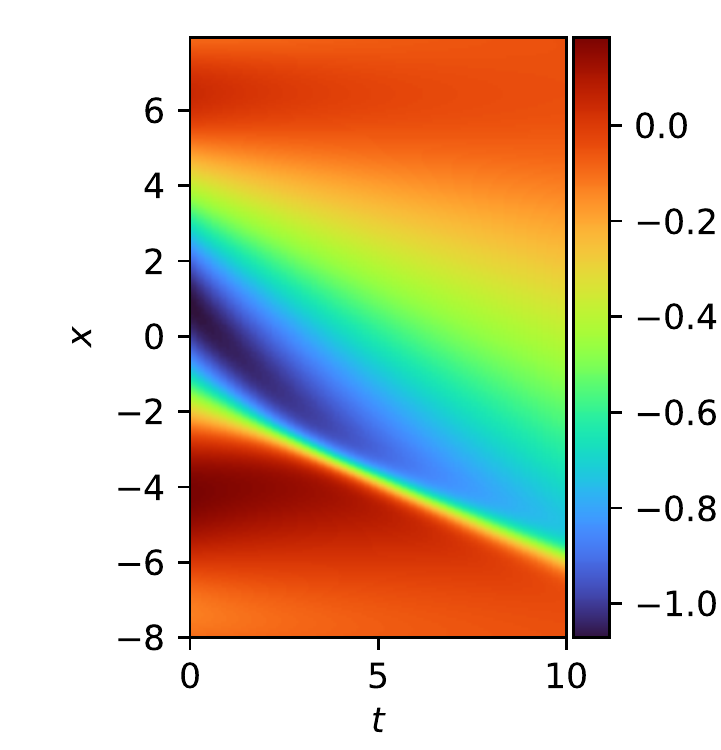}
         \includegraphics[width=0.19\textwidth,clip]{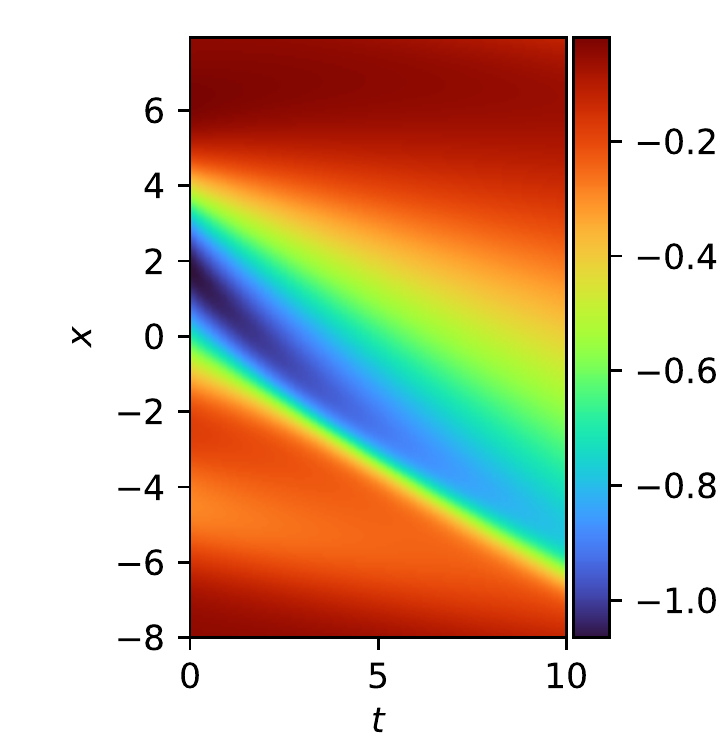}
         \caption{Burgers, training data.}
         \label{fig:burgers_train}
    \end{subfigure}
    \\
    \begin{subfigure}[b]{\textwidth}
         \centering
         \includegraphics[width=0.19\textwidth,clip]{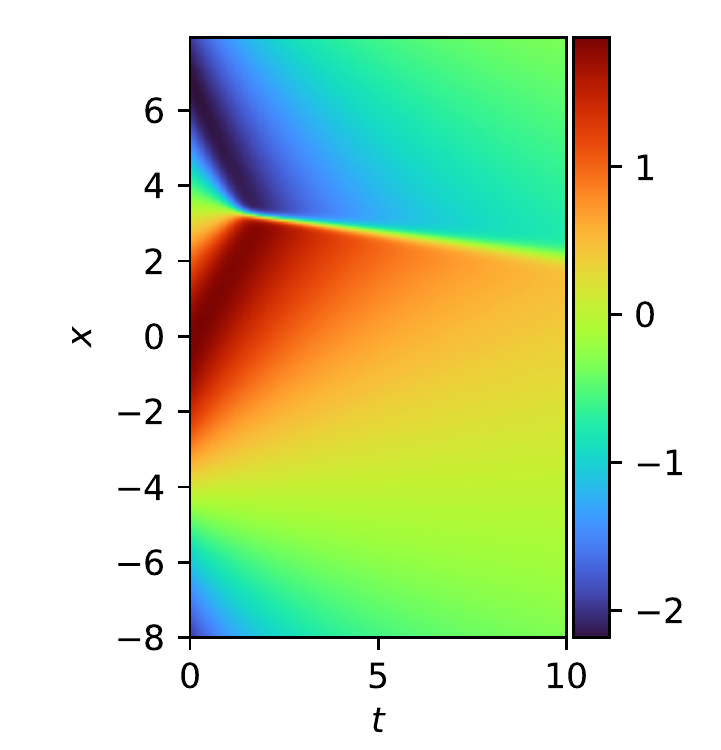}
         \includegraphics[width=0.19\textwidth,clip]{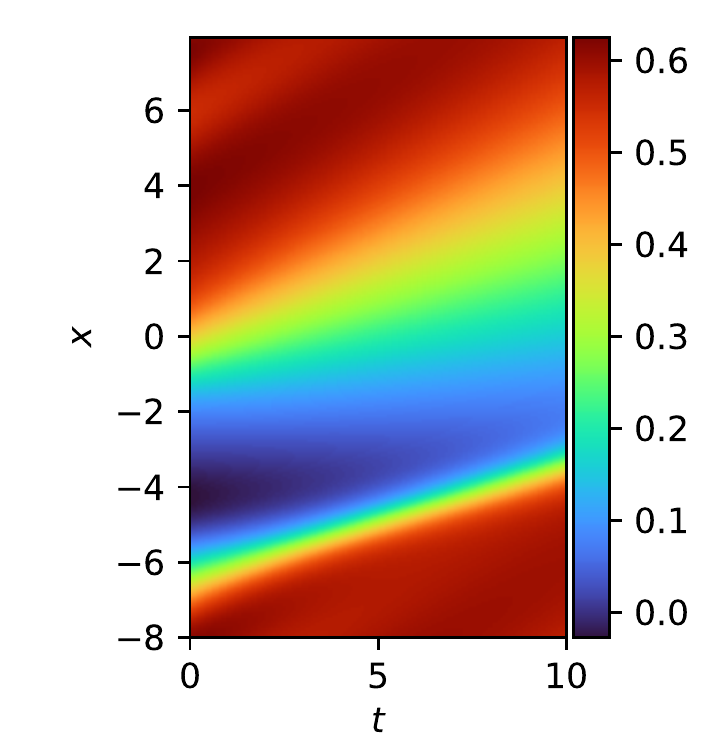}
         \includegraphics[width=0.19\textwidth,clip]{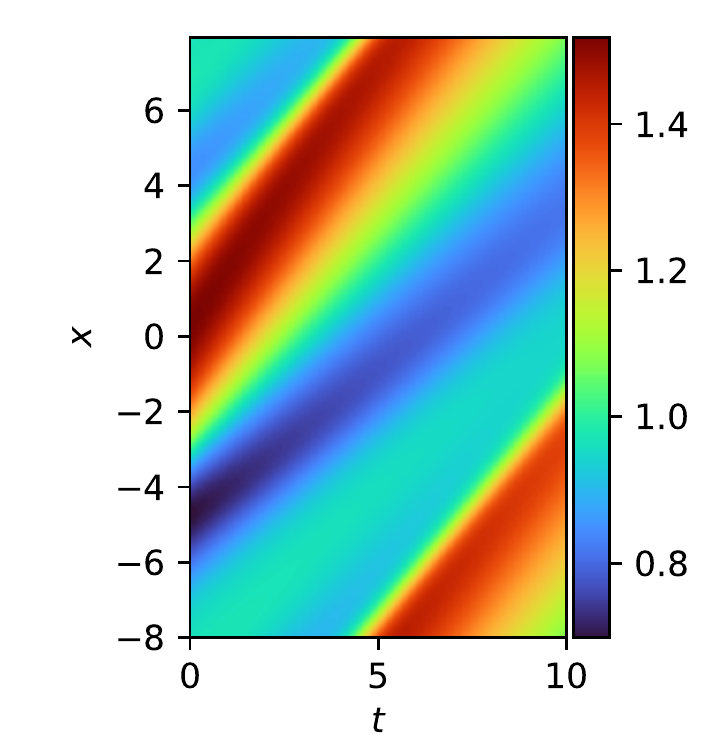}
         \includegraphics[width=0.19\textwidth,clip]{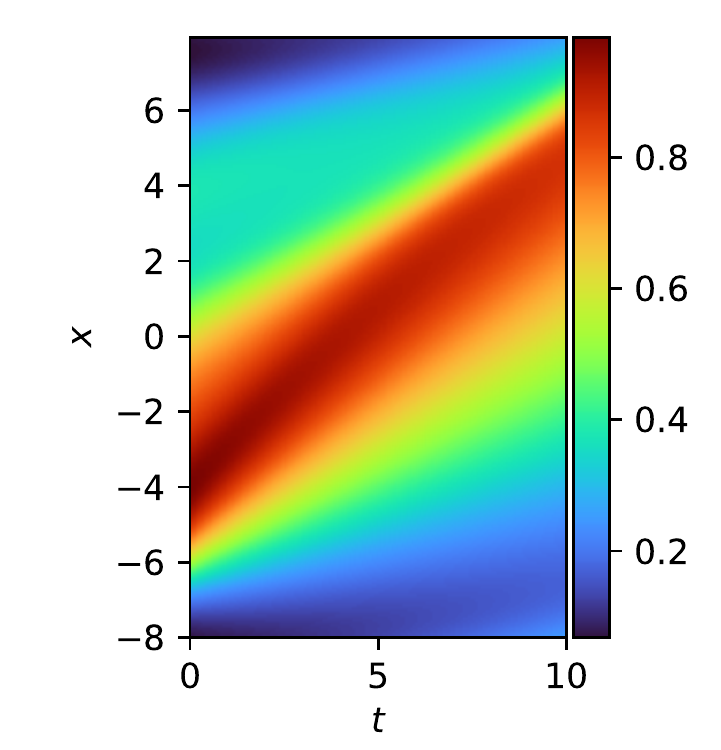}
         \includegraphics[width=0.19\textwidth,clip]{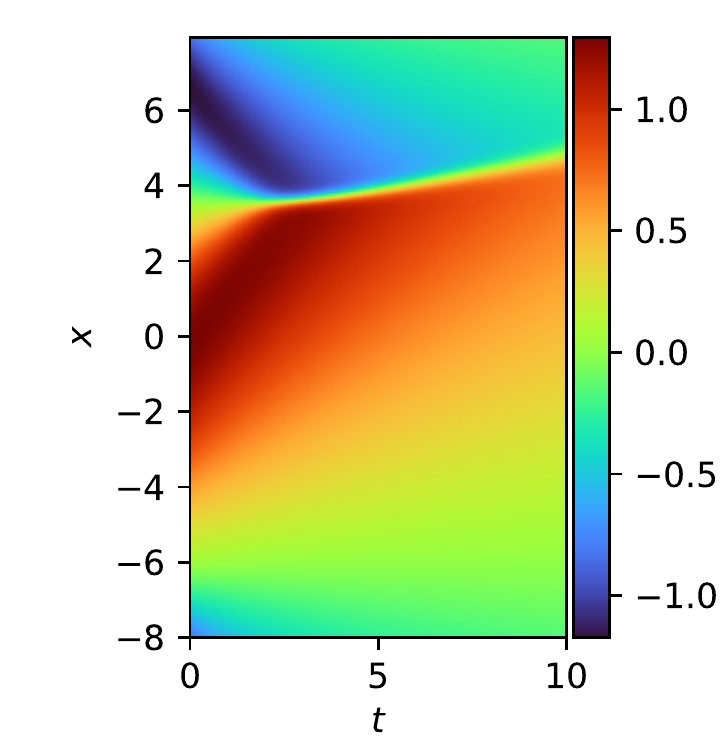}
         \includegraphics[width=0.19\textwidth,clip]{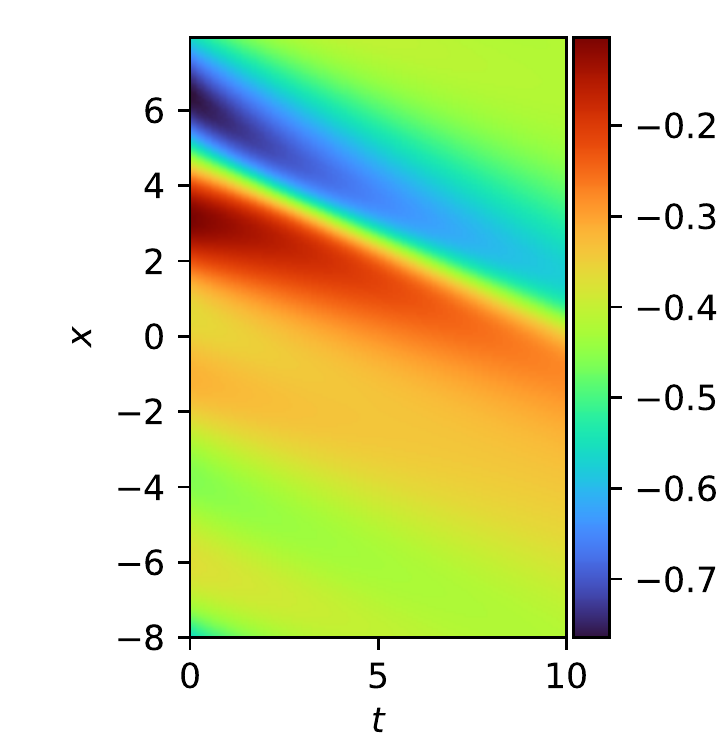}
         \includegraphics[width=0.19\textwidth,clip]{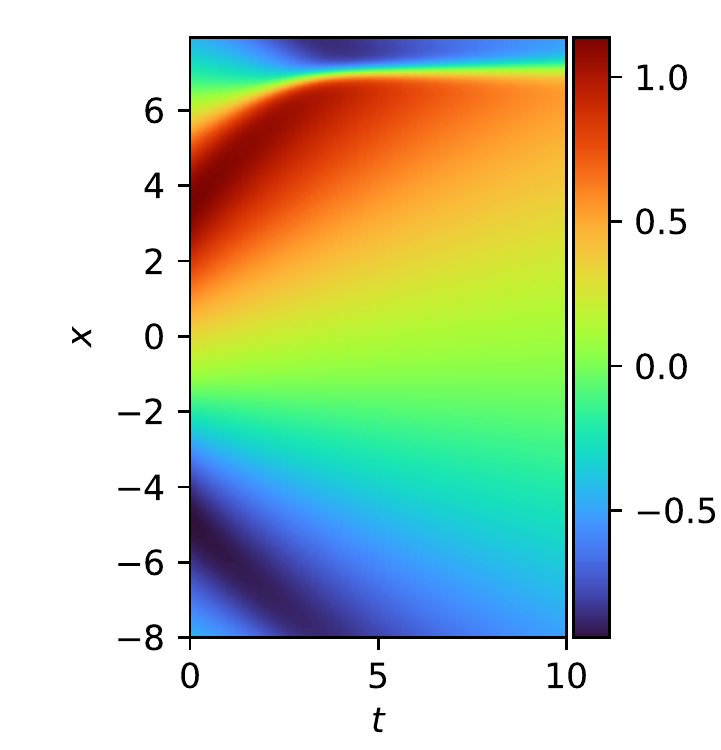}
         \includegraphics[width=0.19\textwidth,clip]{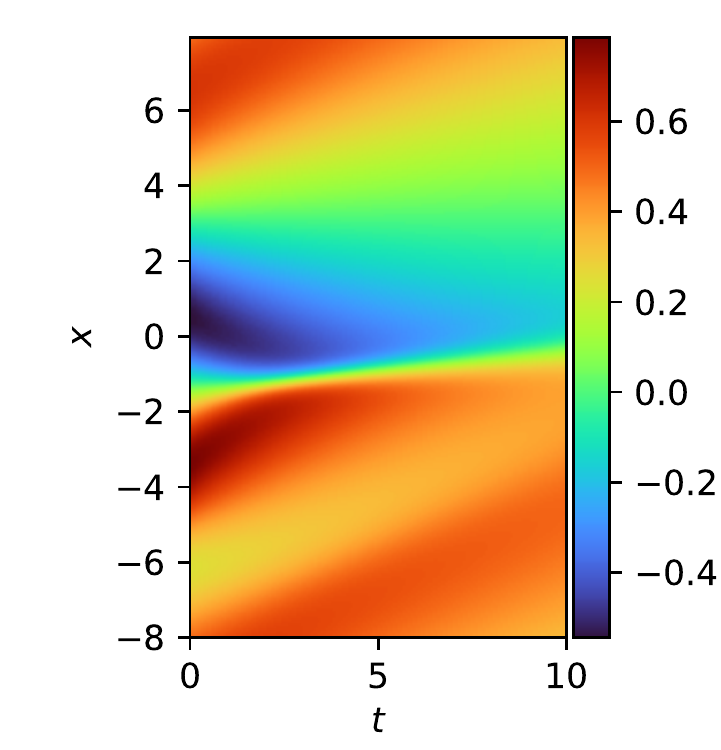}
         \includegraphics[width=0.19\textwidth,clip]{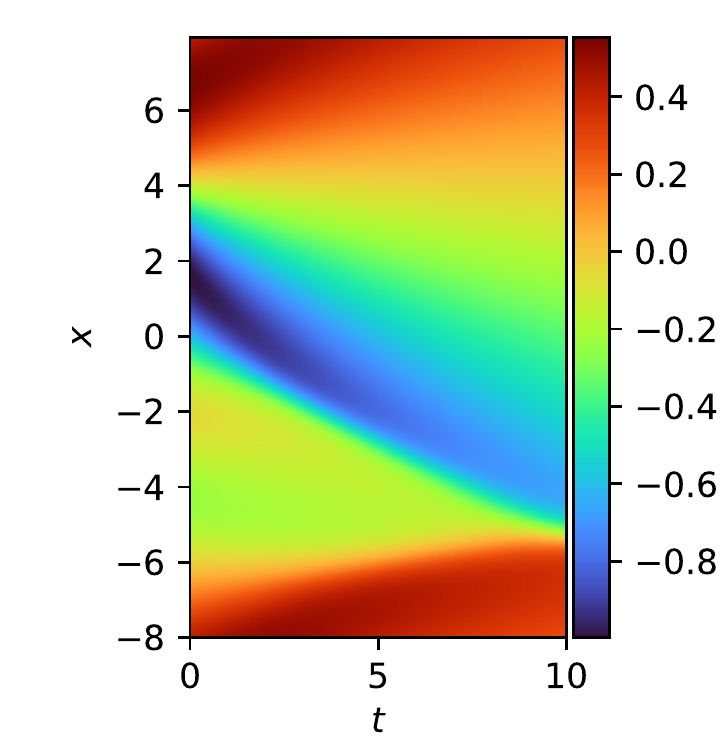}
         \includegraphics[width=0.19\textwidth,clip]{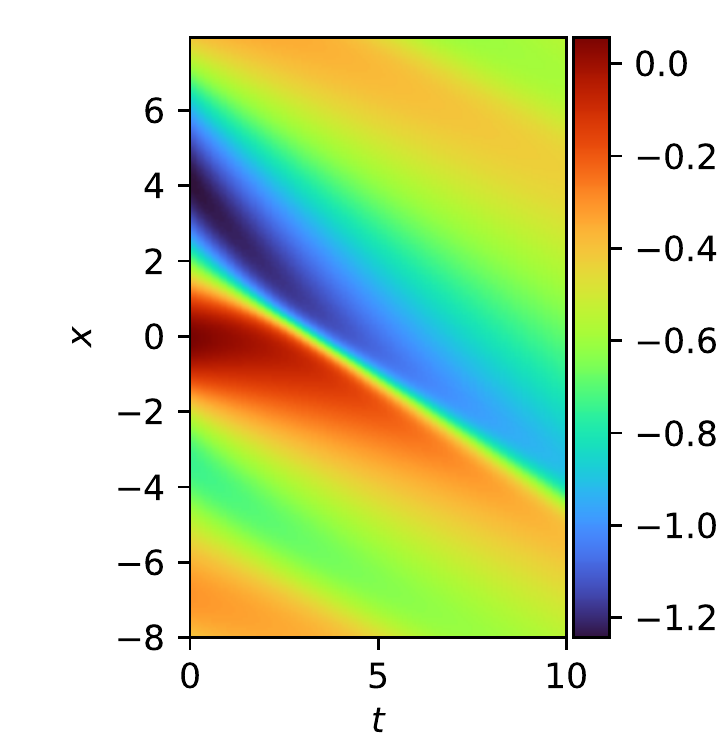}
         \caption{Burgers, testing data.}
         \label{fig:burgers_test}
    \end{subfigure}
    \\
    \begin{subfigure}[b]{\textwidth}
         \centering
         \includegraphics[width=0.19\textwidth,clip]{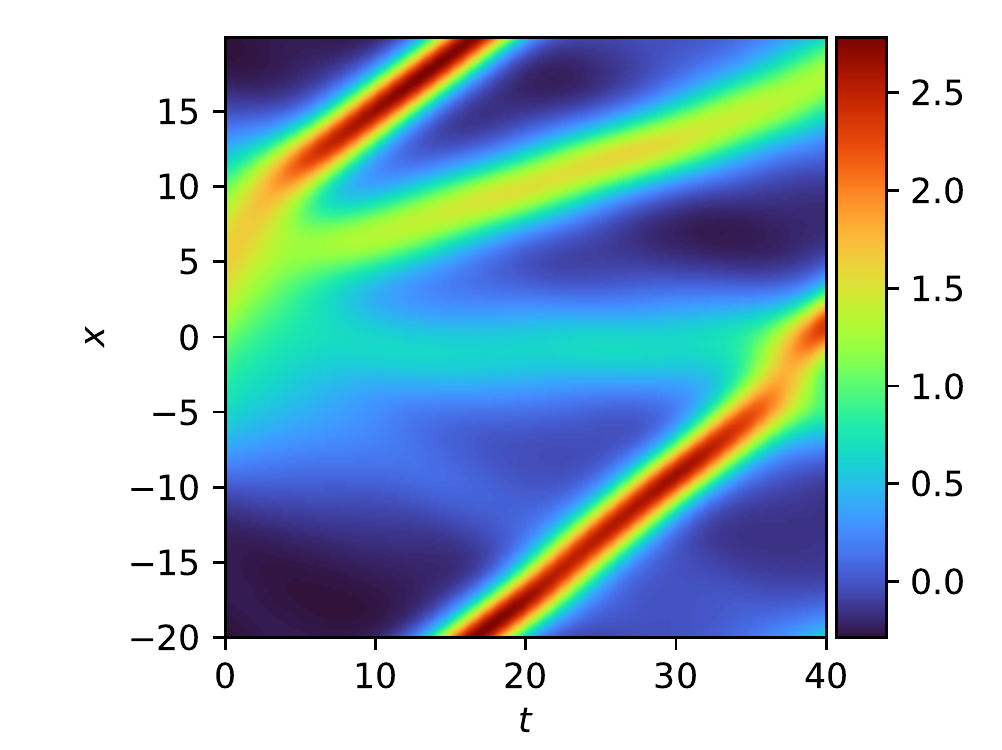}
         \includegraphics[width=0.19\textwidth,clip]{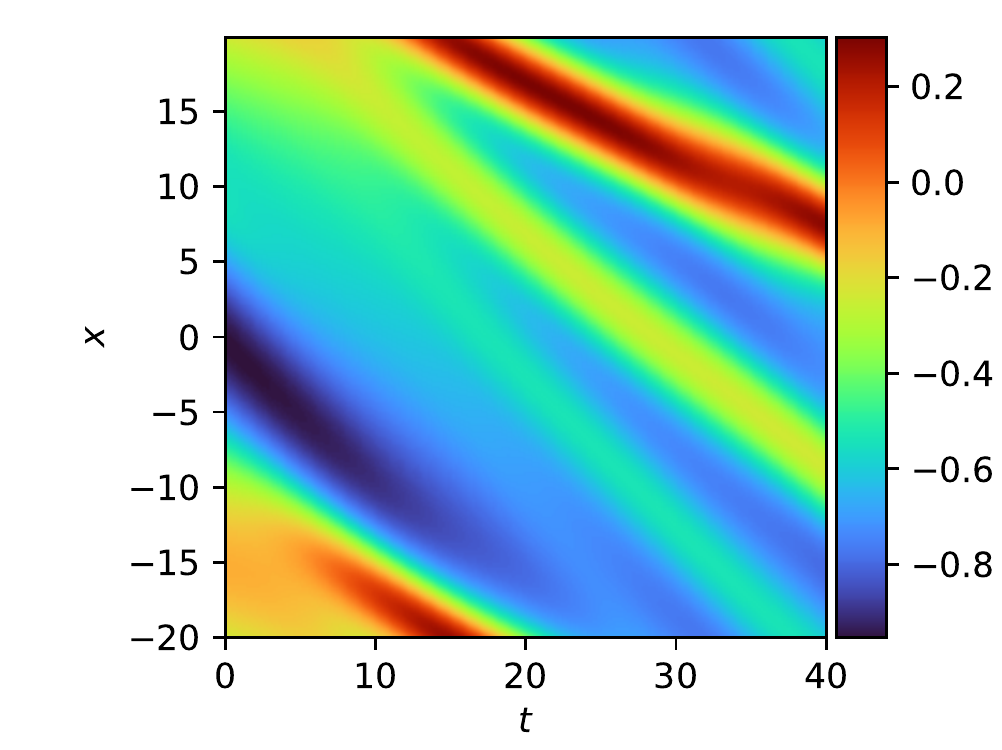}
         \includegraphics[width=0.19\textwidth,clip]{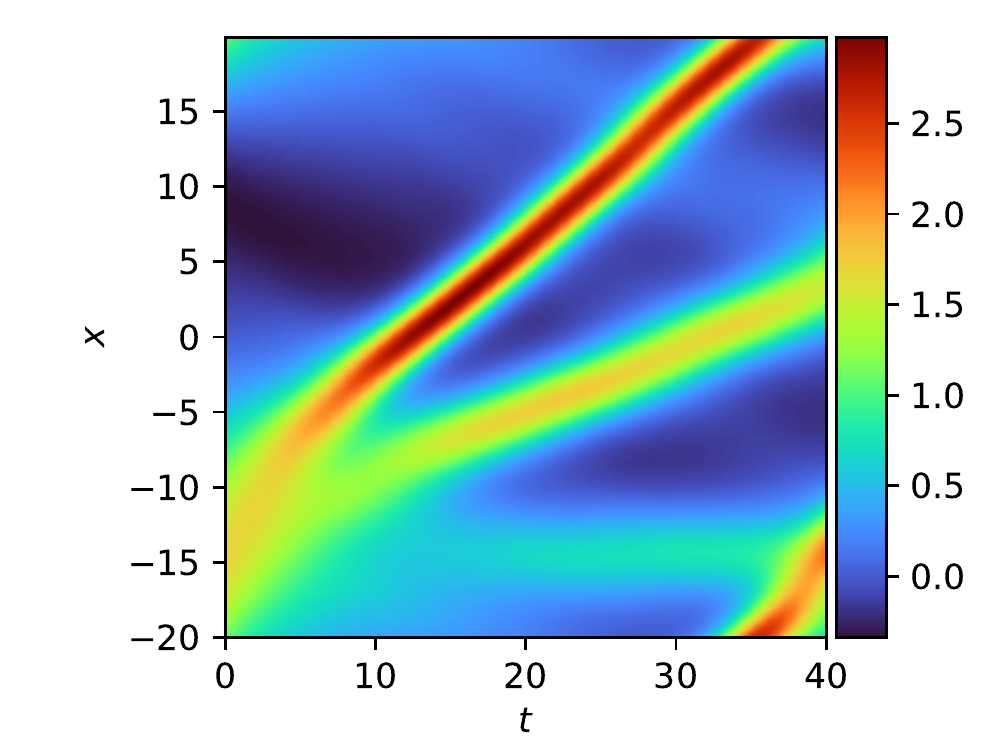}
         \includegraphics[width=0.19\textwidth,clip]{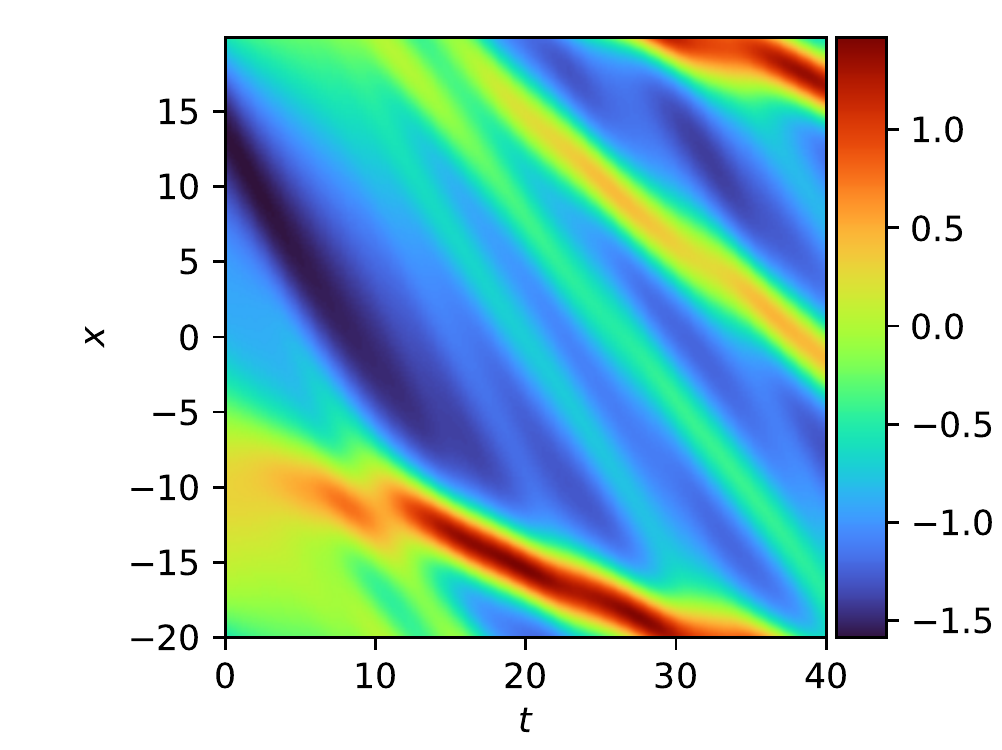}
         \includegraphics[width=0.19\textwidth,clip]{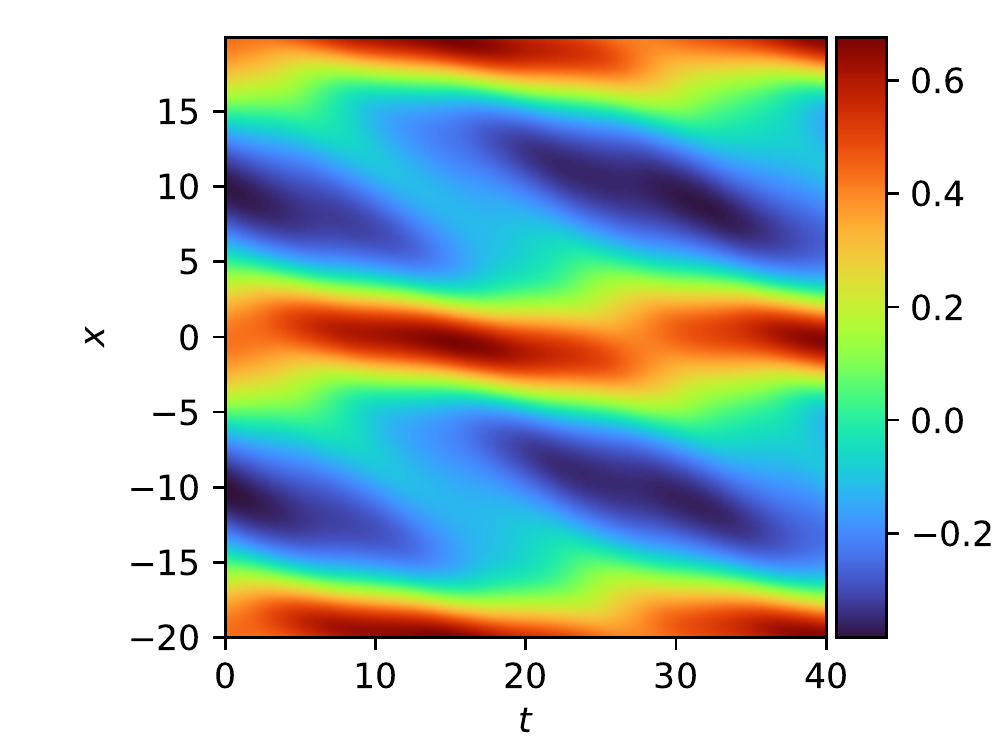}
         \includegraphics[width=0.19\textwidth,clip]{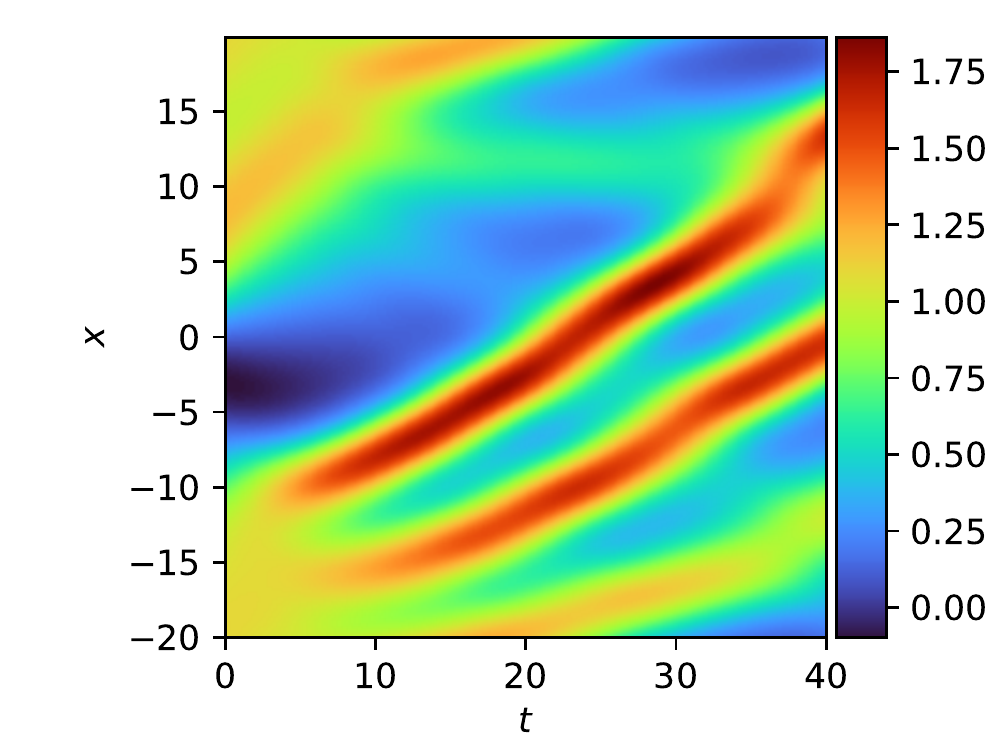}
         \includegraphics[width=0.19\textwidth,clip]{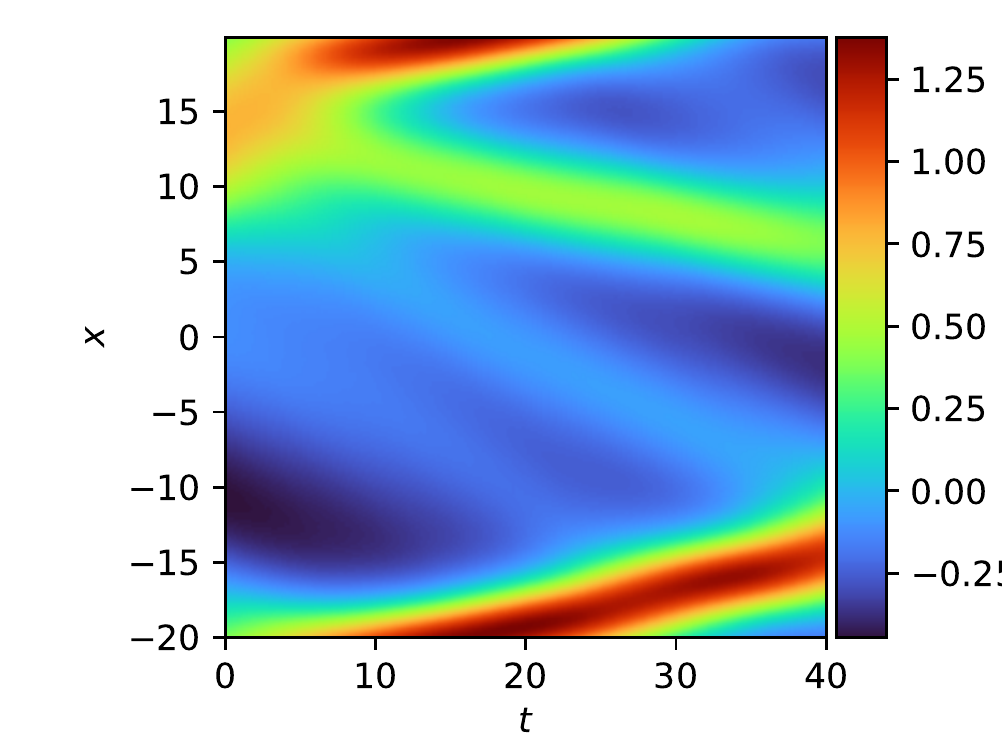}
         \includegraphics[width=0.19\textwidth,clip]{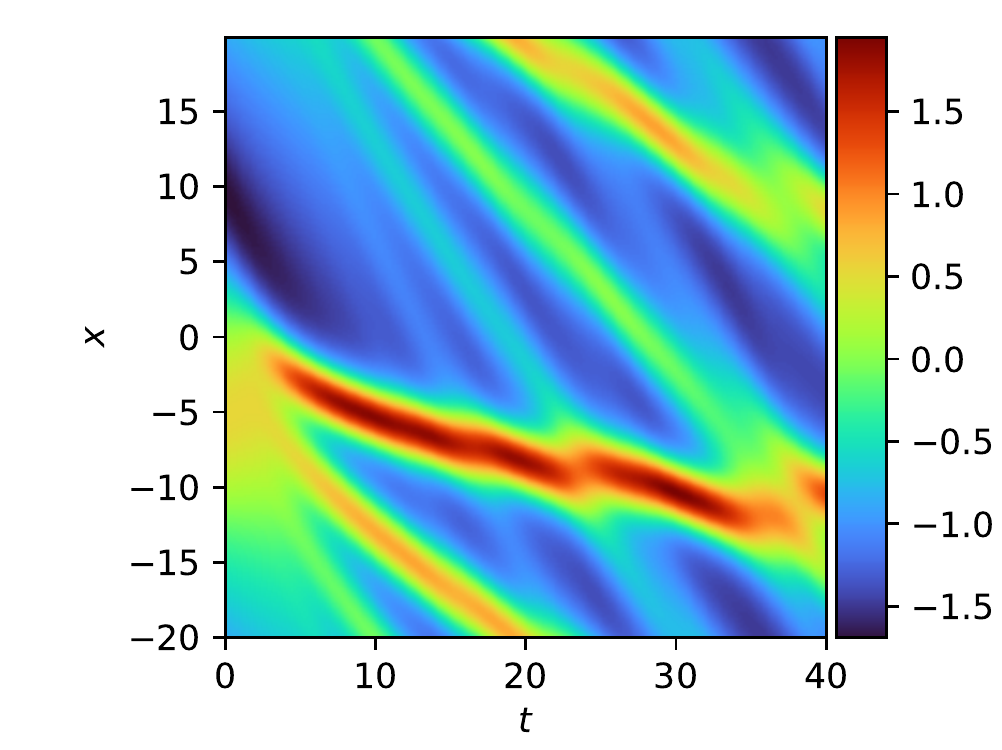}
         \includegraphics[width=0.19\textwidth,clip]{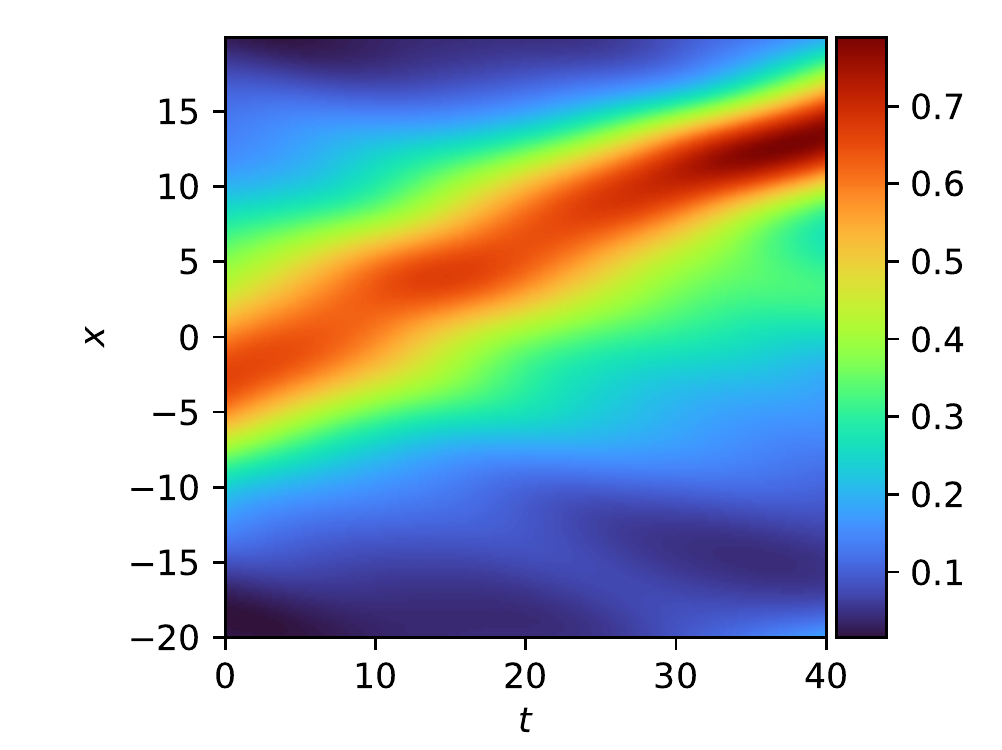}
         \includegraphics[width=0.19\textwidth,clip]{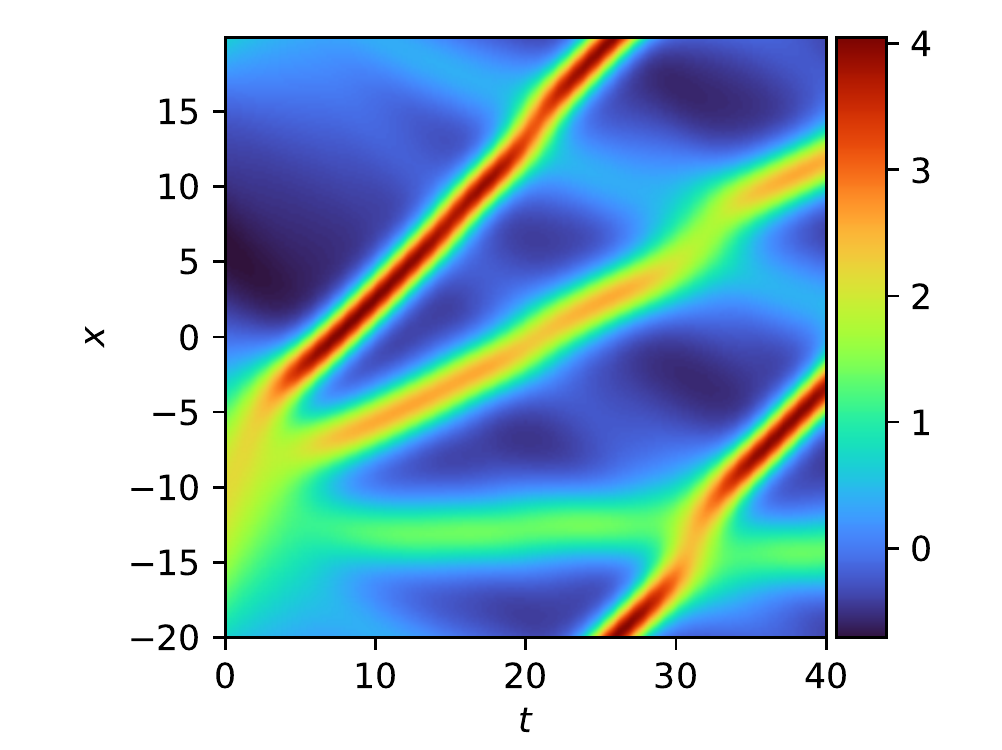}
         \caption{K-dV, training data.}
         \label{fig:kdv_train}
    \end{subfigure}
    \\
    \begin{subfigure}[b]{\textwidth}
         \centering
         \includegraphics[width=0.19\textwidth,clip]{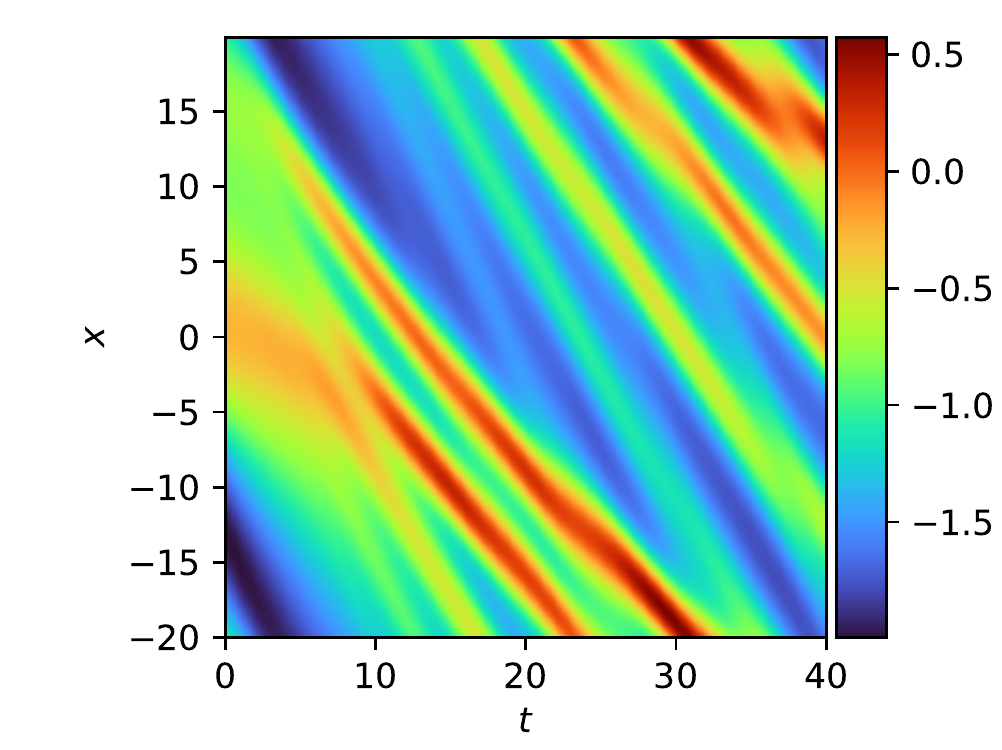}
         \includegraphics[width=0.19\textwidth,clip]{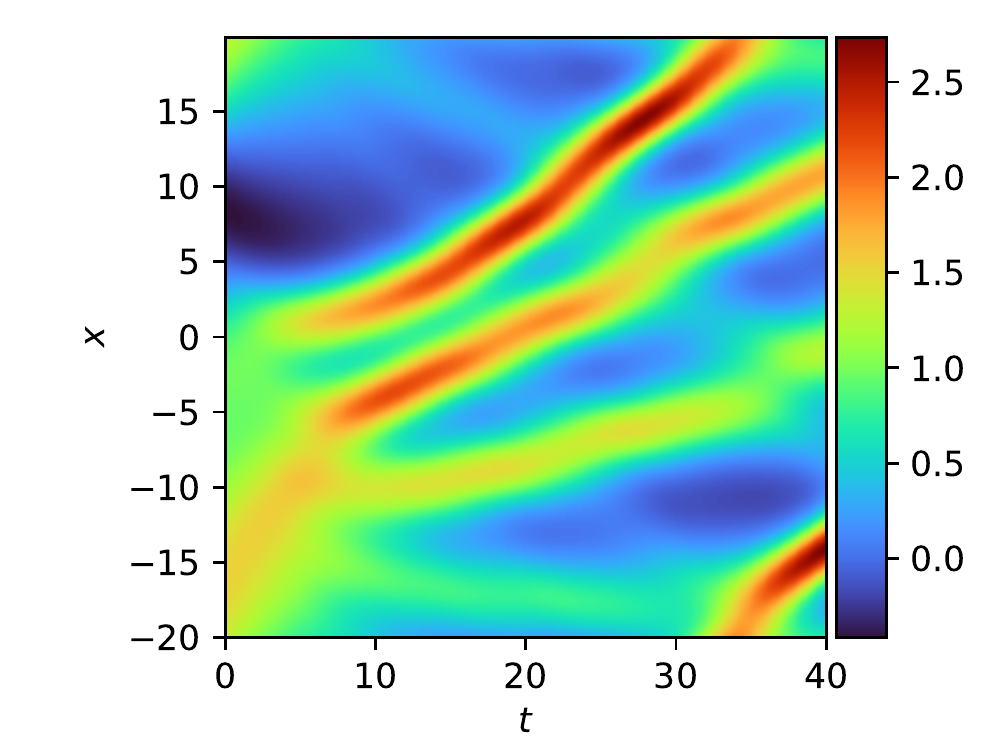}
         \includegraphics[width=0.19\textwidth,clip]{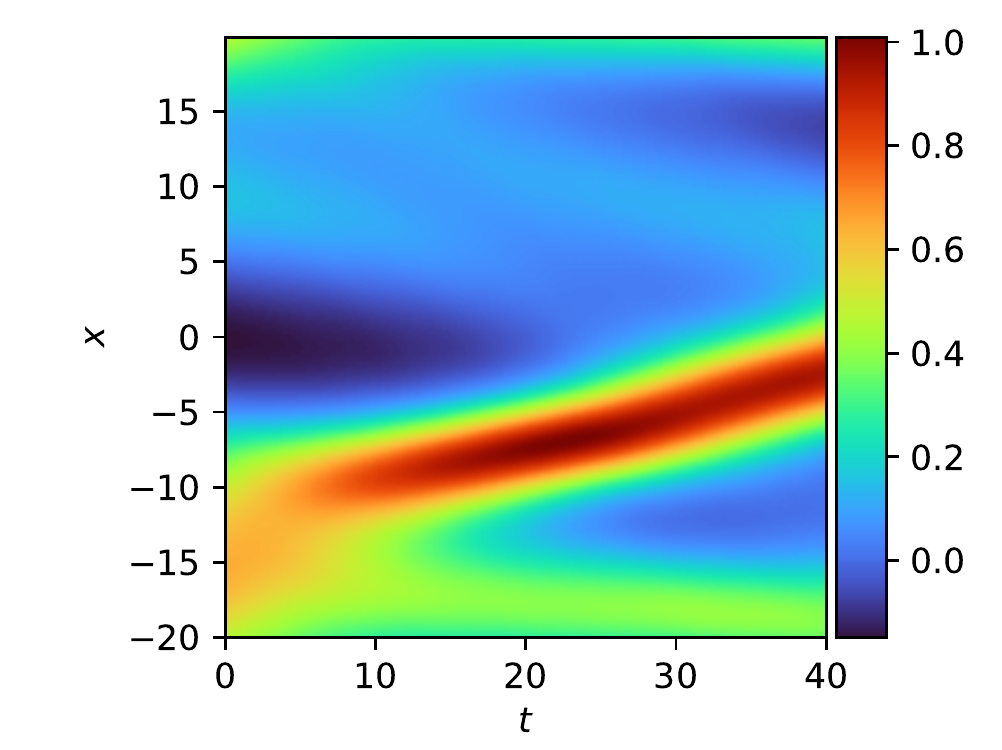}
         \includegraphics[width=0.19\textwidth,clip]{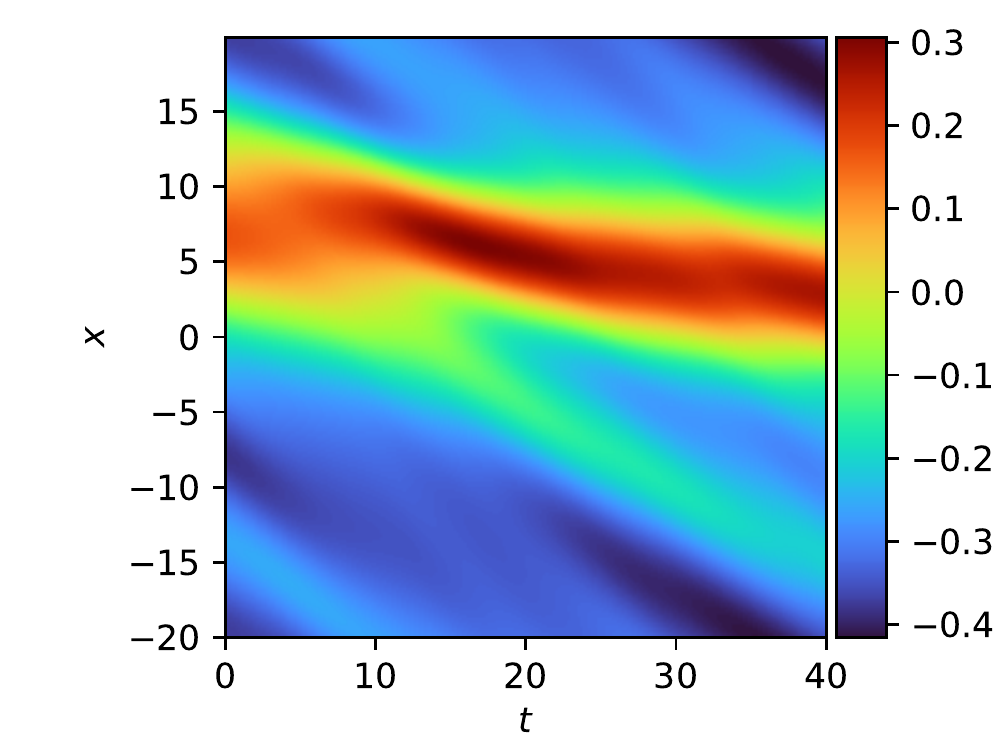}
         \includegraphics[width=0.19\textwidth,clip]{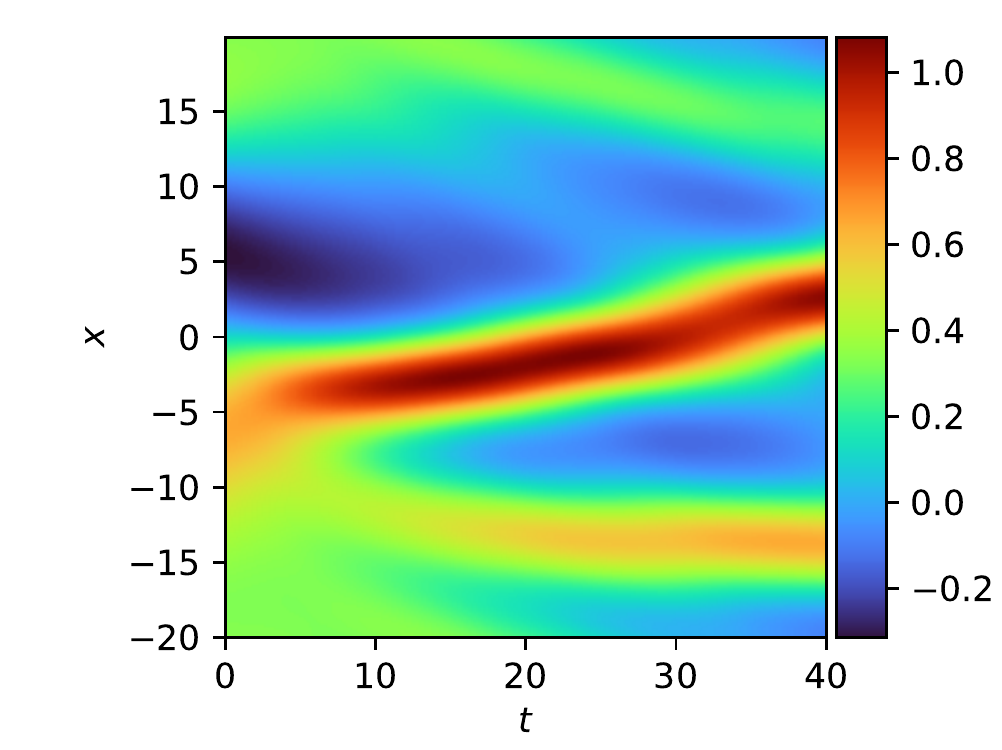}
         \includegraphics[width=0.19\textwidth,clip]{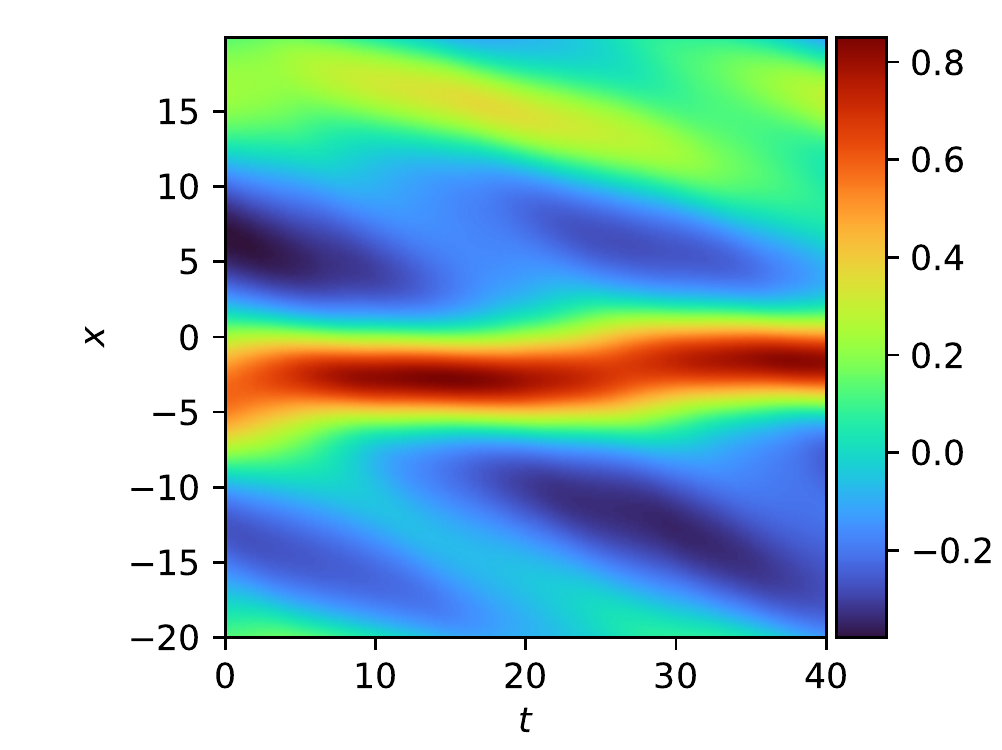}
         \includegraphics[width=0.19\textwidth,clip]{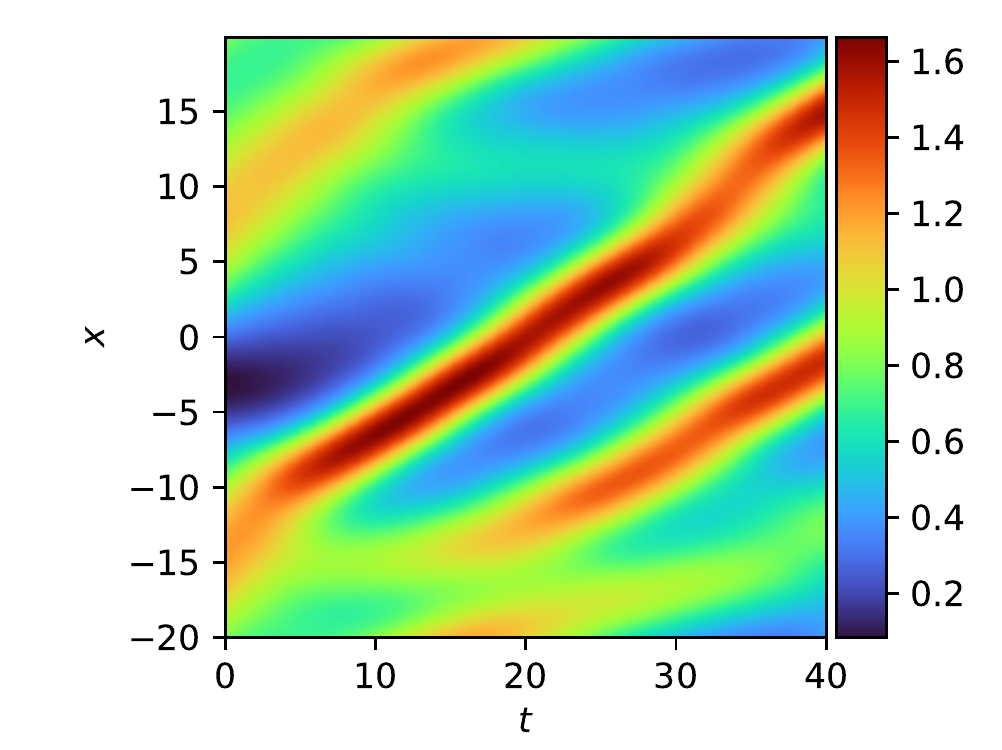}
         \includegraphics[width=0.19\textwidth,clip]{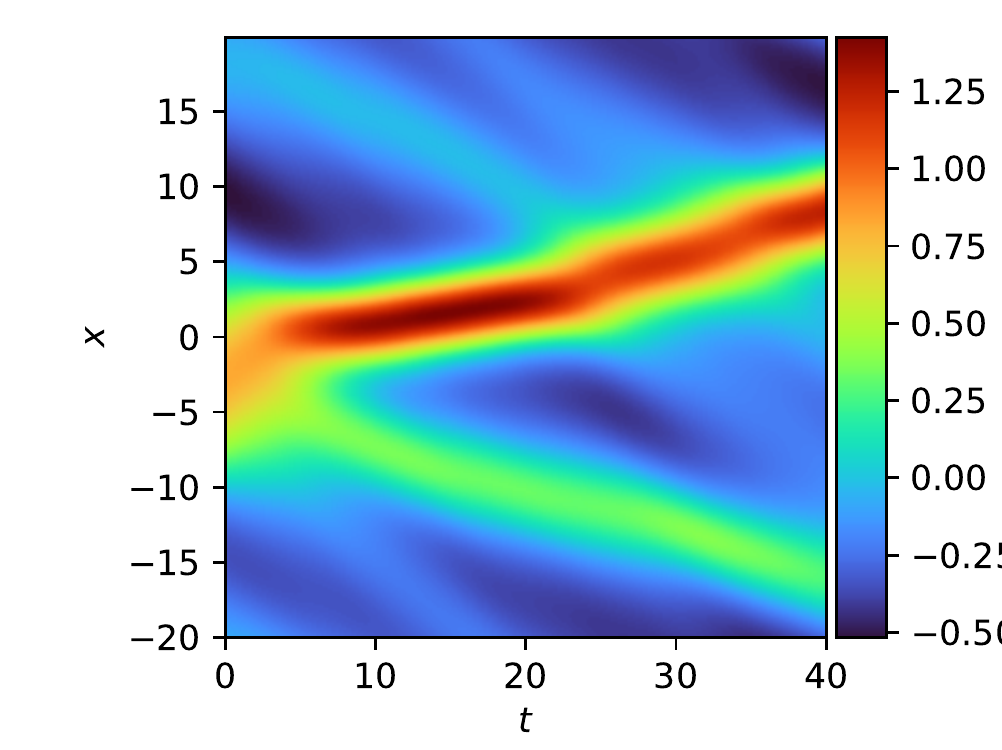}
         \includegraphics[width=0.19\textwidth,clip]{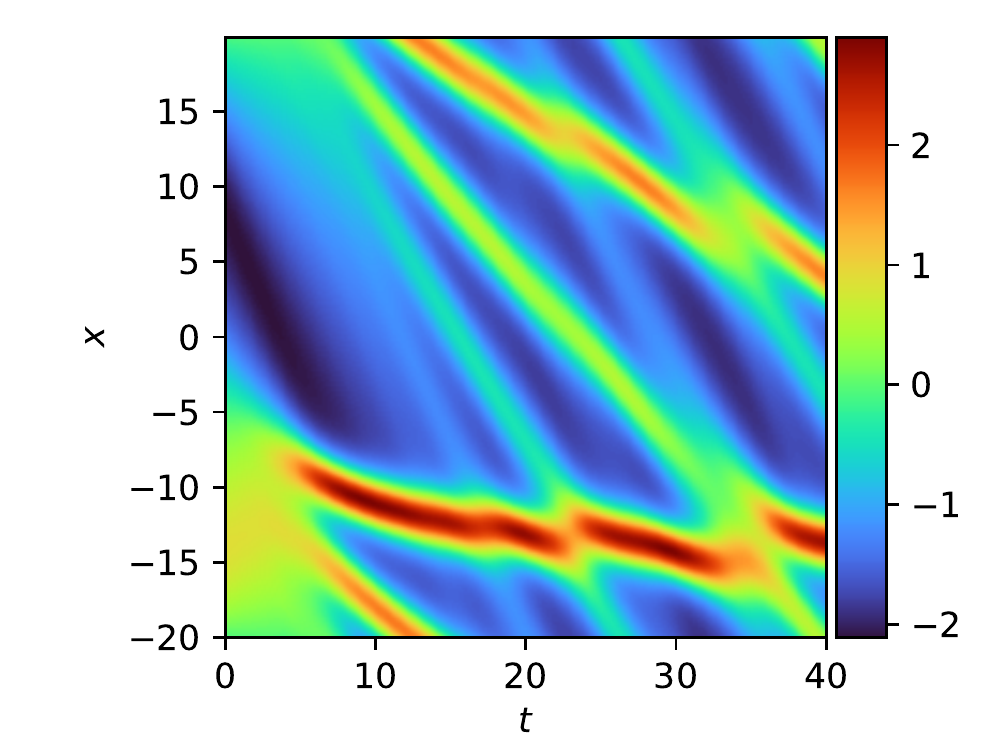}
         \includegraphics[width=0.19\textwidth,clip]{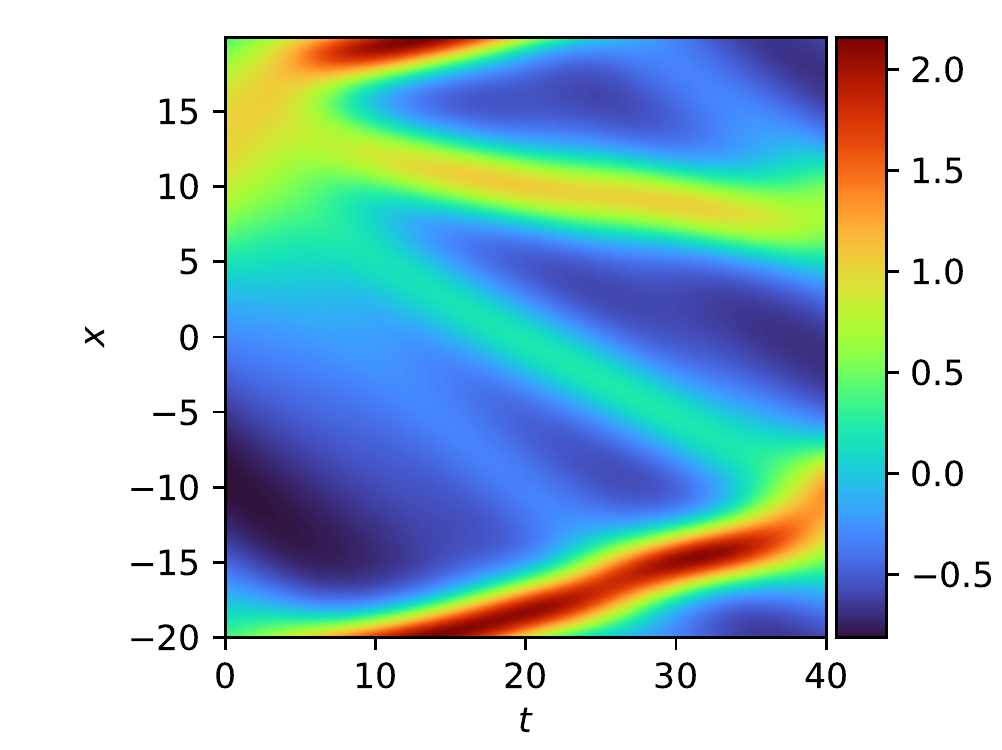}
         \caption{K-dV, testing data.}
         \label{fig:kdv_test}
    \end{subfigure}
    \caption{Datasets used in the experiments}
    \label{fig:data}
\end{figure}

Data are generated using a Python port of Chebfun's \texttt{spin} solver and use the default parameters recommended by the documentation:
For Burgers' equation, we discretize space with 256 spectral elements and a time step of $\Delta t=10^{-4}$.
For the Korteweg-de Vries equation, we discretize space with 512 spectral elements and use a time step of $\Delta t = 10^{-5}$.
Both use a fourth-order Runge-Kutta exponential time differencing scheme \cite{cox2002exponential}.
Following \cite{raissi2018deep}, all simulations were carried out on a rectangular domain 
$\Omega_{st} = \Omega_s \times \Omega_t$ with 
$\Omega_s=[-l_x/2, l_x/2]$ and 
$\Omega_t=0,l_t$ with 
$(l_x, l_t) = (16, 10)$ 
for Burgers' equation and 
$(40, 40)$ 
for the K-dV equation.
Initial conditions were sampled from a Gaussian process with zero mean and periodic kernel $k(x, x') = \exp\left[ \sin^2 \left( \frac{2 \pi (x-x')}{l_x} \right) \right]$,
and periodic spatial boundary conditions are enforced on $\pd{^ku}{x^k}$ for $k=0, \dots, K-1$, where $K$ is the order of the differential equation in space.
Ten solves were created for the training and test sets for each differential equation considered.
Figure \ref{fig:data} shows all of the solves used in this work.

\section{Model architectures and training}
\label{sec:apx:architecture}
For the leaf modules, we use fully-connected neural networks with 6 hidden layers, 64 hidden units per layer, and sine nonlinearities.
The architecture was selected via manual tuning.
We find that smaller models may lack the expressivity to represent the data and larger models incur diminishing returns in terms of accuracy while incurring with higher computational cost.
For the root module, we use a Gaussian process with linear mean function and exponentiated quadratic kernel function.

We train our Bayesian HPM models in three steps.
First, each leaf is trained individually by minimizing its data negative log-likelihood over the neural network parameters and likelihood scale parameter.
Minimization is carried out using the Adam optimizer \cite{kingma2014adam} for $10^4$ iterations per leaf with a learning rate of $10^{-3}$, and all other hyperparameters left at the default values.
No minibatching is used.
Second, a static dataset is generated for the root based on the leaves' current state.
If more than $1024$ points are generated by the leaves, a subset of $1024$ data are randomly subsampled.
The root is trained as a full GP by maximizing the marginal log-likelihood of the data subset with respect to the mean function and kernel hyperparameters.
We train for $2000$ iterations using Adam with a learning rate of $10^{-2}$, leaving all other optimizer hyperparameters at their defaults.
After this initial training for the GP, the model is made sparse by introducing $n_u=128$ inducing inputs at the centers found by $k$-means clustering of the input data; the variational posterior for the inducing outputs was initialized to the full GP's joint posterior distribution at the inducing inputs' locations.
Third, the leaves and sparse GP are trained jointly for $5 \times 10^4$ iterations using the Adam optimizer with learning rate $10^{-3}$, leaving all other optimizer hyperparameters at their default values.

Our implementation uses JAX \cite{bradbury2018jax} to facilitate the gradient computations required of the leaf modules.
Due to the well-known numerical precision challenges associated with working with GP kernel matrices, the model is trained using double precision floating-point arithmetic.
However, due to the limitations of JAX as of version 0.1, this means that we must also train the leaf neural networks in double precision, though this is not intrinsically necessary.

Following the protocol described above, the training time for a Bayesian HPM with $K=2$, $N=4$, and $n_{st}=8192$ is slightly less than 4 hours using a single consumer nVIDIA RTX 2070.
Empirically, scaling is roughly linear in $K$, $N$, and $n_{st}$, though deviations are most pronounced as $n_{st}$ is decreased as the GPU becomes noticeably unsaturated.
Additionally, training time is quite stable due to the lack of control flow.

\section{Physics-informed neural networks}
Here, we provide additional details about the physics-informed neural networks method for solving nonlinear PDEs \cite{raissi2019physics}.
As mentioned in the main text, 
the inputs to the PINNs method are a deterministic differential operator $\tilde f(u, u_x, \dots)$, and boundary operators $\{ b_i(u, u_x, \dots) \}_{i=1}^{n_B}$.
The solution to the differential equation in a $d_{st}$-dimensional spatiotemporal domain $\Omega_{st} \subset \reals^{d_{st}}$ is parameterized by a neural network $\tilde u(x, t; \tilde{\vc \theta})$,
where $\tilde{\vc \theta} \in \Omega_\theta \subseteq \reals^{n_\theta}$ are the network parameters subject to learning.

We assume that the functional form specified by $\tilde u(\cdot)$ is sufficiently expressive that all required derivatives as specified by the inputs to $\tilde f(\cdot)$ and $\{b_i(\cdot)\}_{i=1}^{n_B}$ exist almost everywhere in $\Omega_{st} \times \Omega_\theta$ and are not trivially zero.
For example, a fully-connected network with rectified linear unit nonlinearities would be unsuitable for representing the solution to a PDE that is second-order in space since $\pd{^2 \tilde u}{x^2}$ would be trivially zero almost everywhere.
We use the same architecture for $\tilde u(\cdot)$ as we use for the leaf modules described in Sec.\ \ref{sec:apx:architecture}, ensuring that $\tilde u(\cdot)$ is smooth everywhere in $\Omega_{st} \times \Omega_\theta$.

In order to have $\tilde u(\cdot)$ match the solution to the PDE, the PINNs method seeks to find $\tilde{\vc \theta}$ that minimizes the squared residual associated with the differential operator $\tilde f(\cdot)$ in the domain $\Omega_{st}$ as well as all boundary operators on their relevant portions of the boundary $\partial \Omega_{st}$.
This is accomplished by minimizing the loss function
\begin{equation}
    \begin{aligned}
        L\left(\tilde{\vc \theta}; \mtx X^{(\Omega)}, \mtx X^{(\partial \Omega)} \right) 
        &=
        \frac{1}{n_\Omega} \sum_{i=1}^{n_\Omega} 
        \left( 
            \tilde f \left( \vc v_i^{(\Omega)} \right) 
            - 
            \pd{\tilde u(\vc x_i^{(\Omega)}; \tilde{\vc \theta})}{t}
        \right)^2
        + 
        \sum_{i=1}^{n_B}
        \frac{1}{n_{\partial \Omega,i}}
        \sum_{j=1}^{n_{\partial \Omega,i}}
        b_i \left(\vc v_j^{(\partial \Omega_i)} \right)^2,
    \end{aligned}
    \label{eqn:pinn}
\end{equation}
where $\vc v_i^{(\Omega)} = 
\left(
    \tilde u \left(\vc x_i^{(\Omega)}; \tilde{\vc \theta} \right),
    \pd{\tilde u\left(\vc x_i^{(\Omega)}; \tilde{\vc \theta} \right)}{x},
    \dots
\right)$,
$\vc v_j^{(\partial \Omega_i)} = 
\left(
    \tilde u \left(\vc x_j^{(\partial \Omega_i)}; \tilde{\vc \theta} \right),
    \pd{\tilde u \left(\vc x_j^{(\partial \Omega_i)}; \tilde{\vc \theta} \right)}{x},
    \dots
\right)$,
and
$\mtx X^{(\Omega)} = \left( \vc x_i^{(\Omega)}, \dots, \vc x_{n_\Omega}^{(\Omega)} \right)^\intercal \in \reals^{n_\Omega \times d_{st}}$ and
$\mtx X^{(\partial \Omega_i)} = \left( \vc x_i^{(\partial \Omega_i)}, \dots, \vc x_{n_\Omega}^{(\partial \Omega_i)} \right)^\intercal \in \reals^{n_{\partial \Omega,i} \times d_{st}}$ with $i=1,\dots,n_B$
are collocation points sampled uniformly from the domain $\Omega_{st}$ and each boundary set $\{ \partial \Omega_i \}_{i=1}^{n_B}$ where each respective boundary operator is applied.
We optimize $\tilde{\vc \theta}$ using stochastic optimization, using a batch size of $n_\Omega = 4096$ points for the domain and $n_{\partial \Omega, i} = 256$ for each boundary condition.
We use the Adam optimizer with a cosine-annealed learning rate.
The initial learning rate is $10^{-3}$ and the final learning rate is $10^{-4}$.

\begin{figure}[hbt]
    \centering
    \includegraphics[width=0.5\textwidth]{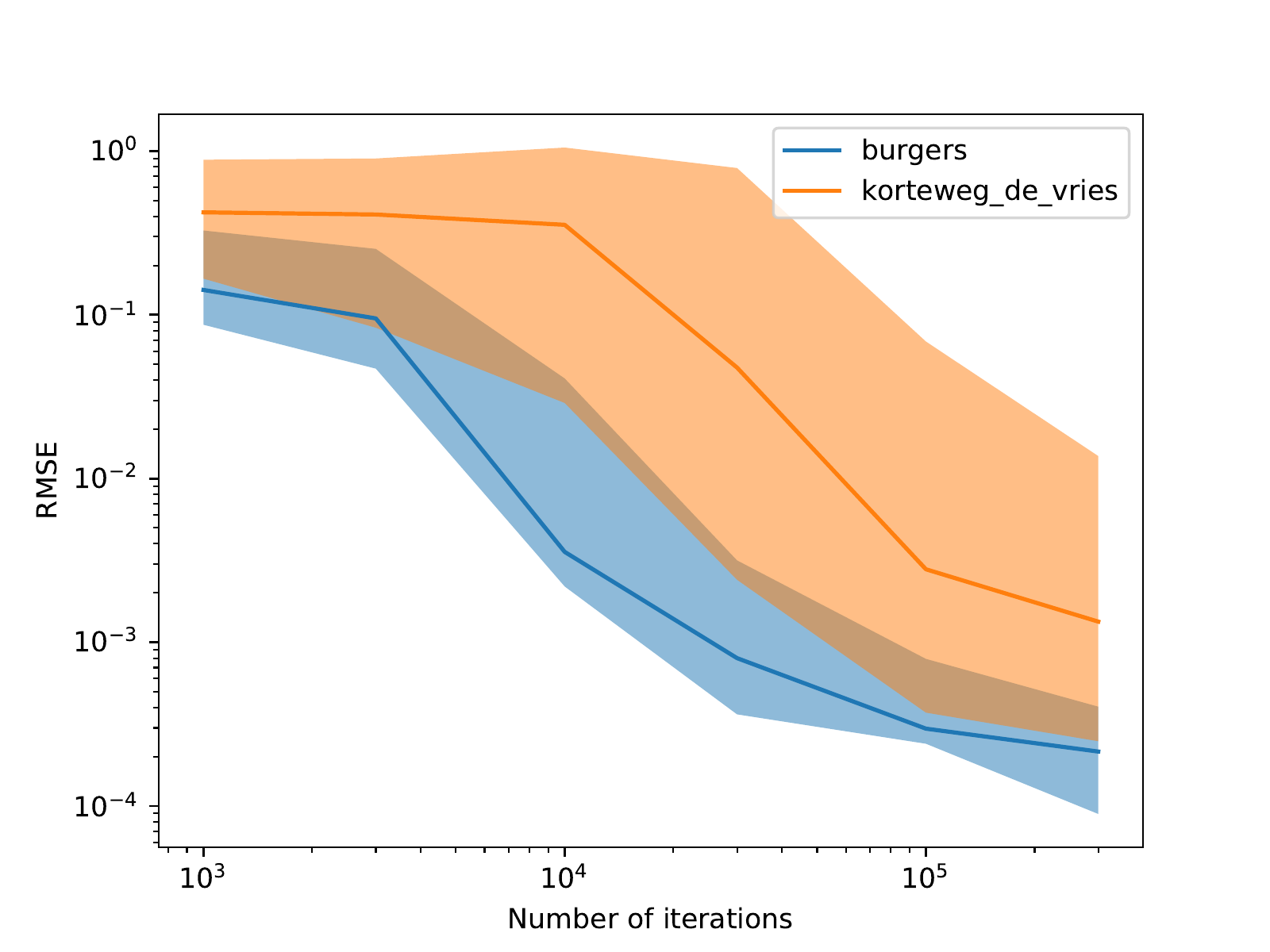}
    \caption{
        Absolute RMSE for solves using the PINNs method with respect to the ground truth data.
        Error bars correspond to empirical 95\% confidence intervals over all $10$ solutions in each equation's training set.
    }
    \label{fig:apx:pinn:iters}
\end{figure}

\begin{figure}[hbt]
    \centering
    \includegraphics[width=\textwidth]{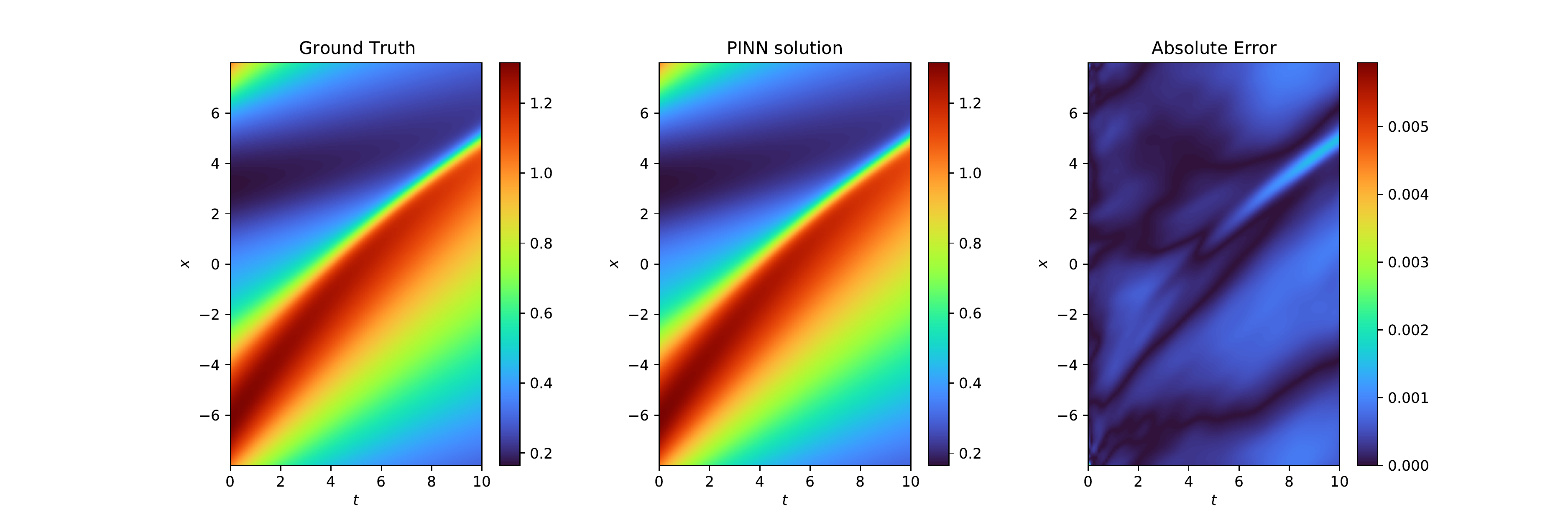}
    \caption{(Burgers' equation) Comparison of a solution using spectral elements (left) and physics-informed neural networks (center).}
    \label{fig:apx:pinn:burgers}
\end{figure}

\begin{figure}[hbt]
    \centering
    \includegraphics[width=\textwidth]{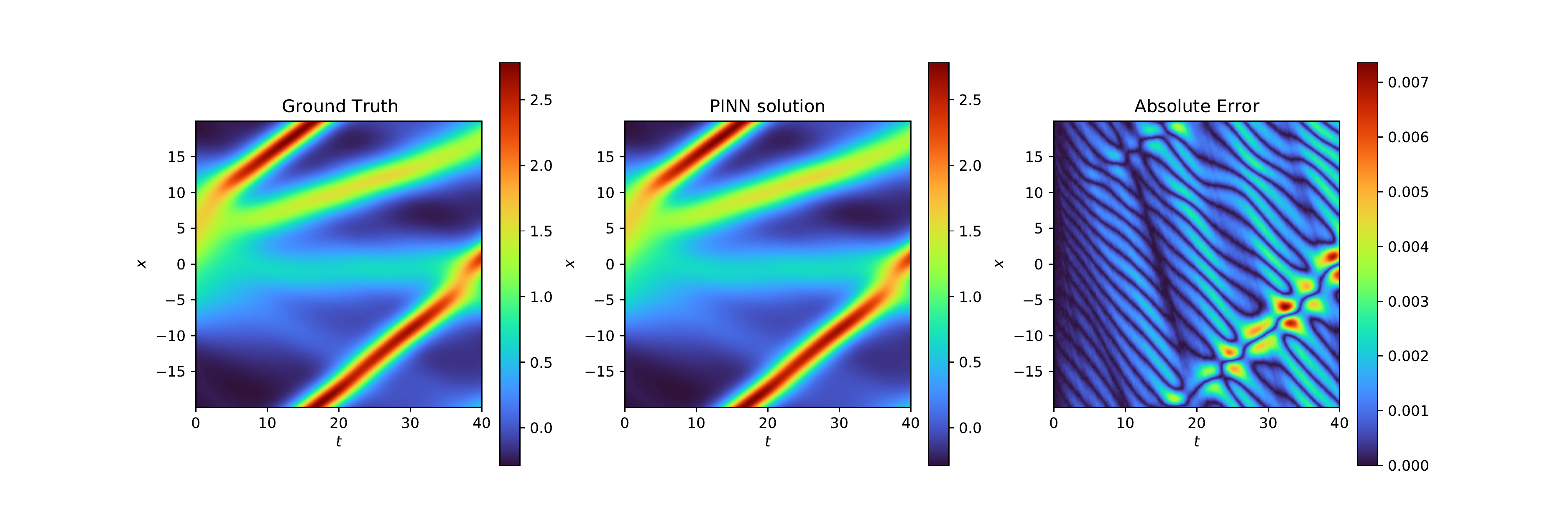}
    \caption{(K-dV equation) Comparison of a solution using spectral elements (left) and physics-informed neural networks (center).}
    \label{fig:apx:pinn:kdv}
\end{figure}

Figure \ref{fig:apx:pinn:iters} shows the convergence of the PINNs method against a ground truth computed with the spectral element method as a function of the number of optimization iterations.
Figures \ref{fig:apx:pinn:burgers} and \ref{fig:apx:pinn:kdv} show example solves with the PINNs algorithm compared to their respective ground truths.
The solution error can be further improved by iterating longer and using more expressive functional forms for $\tilde u(\cdot)$ (e.g.\ using more layers, more units, and other architectural tricks such as skip connections).
In the experiments in the paper, we use $5 \times 10^4$ iterations when solving Burgers' equation and $3 \times 10^5$ iterations when solving the K-dV equation with the PINNs method.

\clearpage
\bibliographystyle{unsrt}

\end{document}